\PassOptionsToPackage{table}{xcolor}
\ifdefined\pdfminorversion\pdfminorversion=7\fi 
\documentclass{article} 
\usepackage[T1]{fontenc}
\usepackage{arxiv_version}
 
\usepackage{microtype}
\usepackage{hyperref}
\usepackage{url}
\usepackage{xurl}
\usepackage{booktabs}
\usepackage{multirow}
\usepackage{lineno}
\usepackage{titlesec}
\usepackage[most]{tcolorbox}
\usepackage{caption}
\usepackage{xspace}

\definecolor{darkblue}{rgb}{0, 0, 0.5}
\hypersetup{
  colorlinks=true,
  citecolor=darkblue,
  linkcolor=darkblue,
  urlcolor=darkblue,
  pdftitle={From Solvers to Research: LLM-Driven Formal Mathematics at the Research Frontier},
  pdfauthor={Eric Jiang; Xiao Liang; Junyi Zhang; Justin Baker; Zhi Zhang; Yingjia Wan; Yubei Li; Jinxi Yu; Mengting Li; Haikang Deng; Alexander Taylor; Rushil Raghavan; Michael Simon; Hyunsik Chae; Jia-Chen Gu; Xinjie He; Kyle Hess; Ethan Ji; Yiwen Kou; Tzu-Han Lin; Jiachen Lu; Jack Mauro; Vigyan Sahai; Zheng Tan; Hengli Li; Yikai Zhang; Kyle Zheng; Lunji Zhu; Dong Liu; Jasdeep Sidhu; Isaac Hair; Eniuce Jun; Ying Nian Wu; Andrea L. Bertozzi; Kai-Wei Chang; Raghu Meka; Matthew Sottile; Nanyun Peng; Amit Sahai; Terence Tao; Wei Wang},
  pdfsubject={Narrative survey and position paper on AI for mathematics, formal proving, research workflows, and human--AI collaboration},
  pdfkeywords={AI for mathematics, formal mathematics, theorem proving, large language models, research agents, human--AI collaboration}
}

\definecolor{revisionblue}{HTML}{000000}
\newif\ifshowrevisions
\showrevisionstrue
\DeclareRobustCommand{\revadd}[1]{%
  {\ifshowrevisions\color{revisionblue}\else\color{black}\fi#1}%
}
\newenvironment{revisionblock}
  {\ifshowrevisions\color{revisionblue}\else\color{black}\fi}
  {}
\usepackage{enumitem}
\usepackage{graphicx}
\usepackage{subcaption}
\usepackage{colortbl}
\usepackage{amsmath}
\usepackage{amssymb}
\usepackage{amsfonts}
\usepackage{amsthm}
\usepackage{nicefrac}
\usepackage[capitalise,noabbrev]{cleveref}
\usepackage{mathtools}
\usepackage{siunitx}    
\usepackage{makecell,threeparttable}
\usepackage{pifont}
\usepackage{wrapfig}
\usepackage{float}
\usepackage{needspace}
\usepackage{rotating}
\usepackage{tabularx}
\usepackage{longtable}
\usepackage{etoc}
\usepackage{listings}
\usepackage{natbib}
\usepackage{soul}
\usepackage{CJKutf8}

\usepackage{tikz}
\usetikzlibrary{shadows, arrows.meta, positioning, shapes.geometric, calc}
\usepackage[edges]{forest}

\usepackage{fvextra}

\DefineVerbatimEnvironment{LeanCode}{Verbatim}{
  fontsize=\footnotesize,
  breaklines=true,
  breakanywhere=true,
  baselinestretch=1.0
}

\crefname{section}{Section}{Sections}
\Crefname{section}{Section}{Sections}
\crefrangelabelformat{section}{#3#1--#4#2}

\crefname{subsection}{\S}{\S}
\Crefname{subsection}{\S}{\S}

\newcommand{\revisiondate}{1 Nov 2026}
\renewcommand{\today}{\revisiondate}
\newcommand{\literaturecutoff}{20 October 2026}

\definecolor{lightcoral}{rgb}{0.94, 0.5, 0.5}
\definecolor{darkpastelgreen}{rgb}{0.01, 0.75, 0.24}
\definecolor{hidden-red}{RGB}{205, 44, 36}
\definecolor{hidden-blue}{RGB}{194,232,247}
\definecolor{hidden-orange}{RGB}{243,202,120}
\definecolor{hidden-green}{RGB}{34,139,34}
\definecolor{hidden-pink}{RGB}{255,245,247}
\definecolor{hidden-black}{RGB}{20,68,106}
\definecolor{purple}{RGB}{144,153,196}
\definecolor{yellow}{RGB}{255,228,123}
\definecolor{hidden-yellow}{RGB}{255,248,203}
\definecolor{tkcolor}{RGB}{224,223,255}
\definecolor{myred}{RGB}{247,226,231}
\definecolor{myblue}{RGB}{216,226,234}
\definecolor{myyellow}{RGB}{252,238,221}
\definecolor{mypurple}{RGB}{233,229,241}
\definecolor{mygreen}{RGB}{204,231,207}

\theoremstyle{plain}

\theoremstyle{definition}

\theoremstyle{remark}

\title{\revadd{From Solvers to Research: LLM-Driven Formal Mathematics at the Research Frontier}}

\author{
Eric~Jiang$^{1,*}$,
Xiao~Liang$^{1,*}$,
Junyi~Zhang$^{1}$,
Justin~Baker$^{1}$,
Zhi~Zhang$^{1}$,
Yingjia~Wan$^{1}$,
Yubei~Li$^{1}$,
Jinxi~Yu$^{1}$,
Mengting~Li$^{1}$,
Haikang~Deng$^{1}$, 
Alexander~Taylor$^{1}$,
Rushil~Raghavan$^{1}$,
Michael~Simon$^{1}$,
Hyunsik~Chae$^{1}$,
Jia-Chen~Gu$^{1}$,
Xinjie~He$^{1}$,
Kyle~Hess$^{1}$,
Ethan~Ji$^{1}$,
Yiwen~Kou$^{1}$,
Tzu-Han~Lin$^{1}$,
Jiachen~Lu$^{1}$,
Jack~Mauro$^{1}$,
Vigyan~Sahai$^{1}$,
Zheng~Tan$^{1}$,
Hengli~Li$^{1}$,
Yikai~Zhang$^{1}$,
Kyle~Zheng$^{1}$,
Lunji~Zhu$^{1}$,
Dong~Liu$^{1}$,
Jasdeep~Sidhu$^{1}$,
Isaac~Hair$^{1}$,
Eniuce~Jun$^{1}$,
Ying~Nian~Wu$^{1}$,
Andrea~L.~Bertozzi$^{1, \dagger}$,
Kai-Wei~Chang$^{1, \dagger}$, 
Raghu~Meka$^{1, \dagger}$, 
Matthew~Sottile$^{2, \dagger}$, 
Nanyun~Peng$^{1, \dagger}$,
Amit~Sahai$^{1, \dagger}$,
Terence~Tao$^{1, \dagger}$,
Wei~Wang$^{1, \dagger}$ \\
[1em]
$^{1}$University of California, Los Angeles \\
$^{2}$Lawrence Livermore National Laboratory \\
}

\begin{document}

\maketitle

\begingroup
  \makeatletter
  \renewcommand\@makefnmark{}%
  \makeatother
  \footnotetext[0]{%
    $^{*}$Equal contribution. \\
    \space $^{\dagger}$ Senior authors served as equal advisors. 
  }%
\endgroup

\begin{abstract}
Large language models have made rapid progress in generating machine-checked proofs for well-defined mathematical problems. Extending this progress to research mathematics requires more than solving isolated theorems. Research involves choosing questions, developing definitions and conjectures, connecting results, and combining formal proofs with expert judgment. This survey examines how LLM-driven systems can support these broader activities. We review formal-mathematics tasks, datasets, and autoformalization, and organize proof-generation methods into three overlapping areas: training strategies, inference-time search and adaptation, and agentic workflows. We then examine benchmark results, selected research projects, and platforms for collaborative mathematical work. A central distinction runs throughout the survey: checking a formal proof does not by itself establish that it captures the intended problem or makes a new mathematical contribution. We therefore distinguish theorem benchmarks, repository tasks, formalization projects, and research claims, and discuss the evidence needed to assess each. Finally, we identify five priorities for progress: better data and evaluation, representations of mathematical relationships, conjecture and concept generation, reliable tool integration, and human--AI collaboration. Together, these perspectives connect advances in proof generation to the practical demands of mathematical research.

\vspace{-0.5em}
\begin{center}
\raisebox{-0.2em} {\includegraphics[height=1.2em]{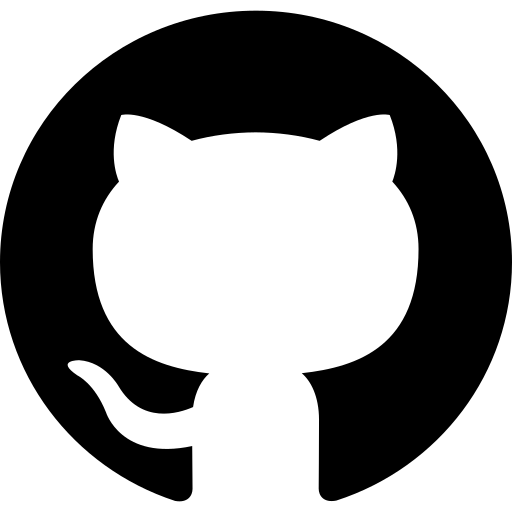}}\hspace{0.3em}
\href{https://github.com/ericjiang18/Awesome-Formal-Mathematics/tree/main}{\texttt{Collection of Resources}}
\end{center}

\end{abstract}

\section{Introduction}
\label{sec:xiao_intro}

Mathematical research is a process of choosing questions, developing definitions and conjectures, finding useful arguments, and communicating results that others can assess and build on. Artificial intelligence for mathematics (AI4Math) increasingly supports this process through systems that combine language models, symbolic reasoning, formal proof assistants, and interactive research tools~\citep{lu2024ai4math,JuDong2026AI4MathSurvey}. As these systems take on broader roles, mathematics becomes a demanding setting for studying how learned systems search, use tools, retain knowledge, and collaborate with experts. This expansion also sharpens a central question: what evidence is needed to assess a system's contribution to mathematical research?

Formal mathematics offers a concrete starting point. Interactive theorem provers represent definitions, statements, and proofs in languages whose derivations can be checked within an explicit logical environment~\citep{moura2021lean4}. Learning-guided theorem proving uses neural models to select premises, propose proof steps, and direct search within this structure~\citep{yang2019coqgym,bansal2019holist,lample2022hypertree}. Large language models extend these methods by translating between informal and formal mathematics, generating proof sketches and complete proofs, and adapting to feedback from the prover~\citep{weng2025autoformalization,xin2024deepseekproverv15,ren2025deepseekproverv2}. Formal environments therefore provide both feedback for learning and a means of checking generated artifacts. Coupling learned guidance with formal checking allows systems to explore large spaces of candidate arguments while retaining an explicit criterion for proof acceptance.

\begin{figure}[t]
    \centering
    \includegraphics[width=0.98\linewidth]{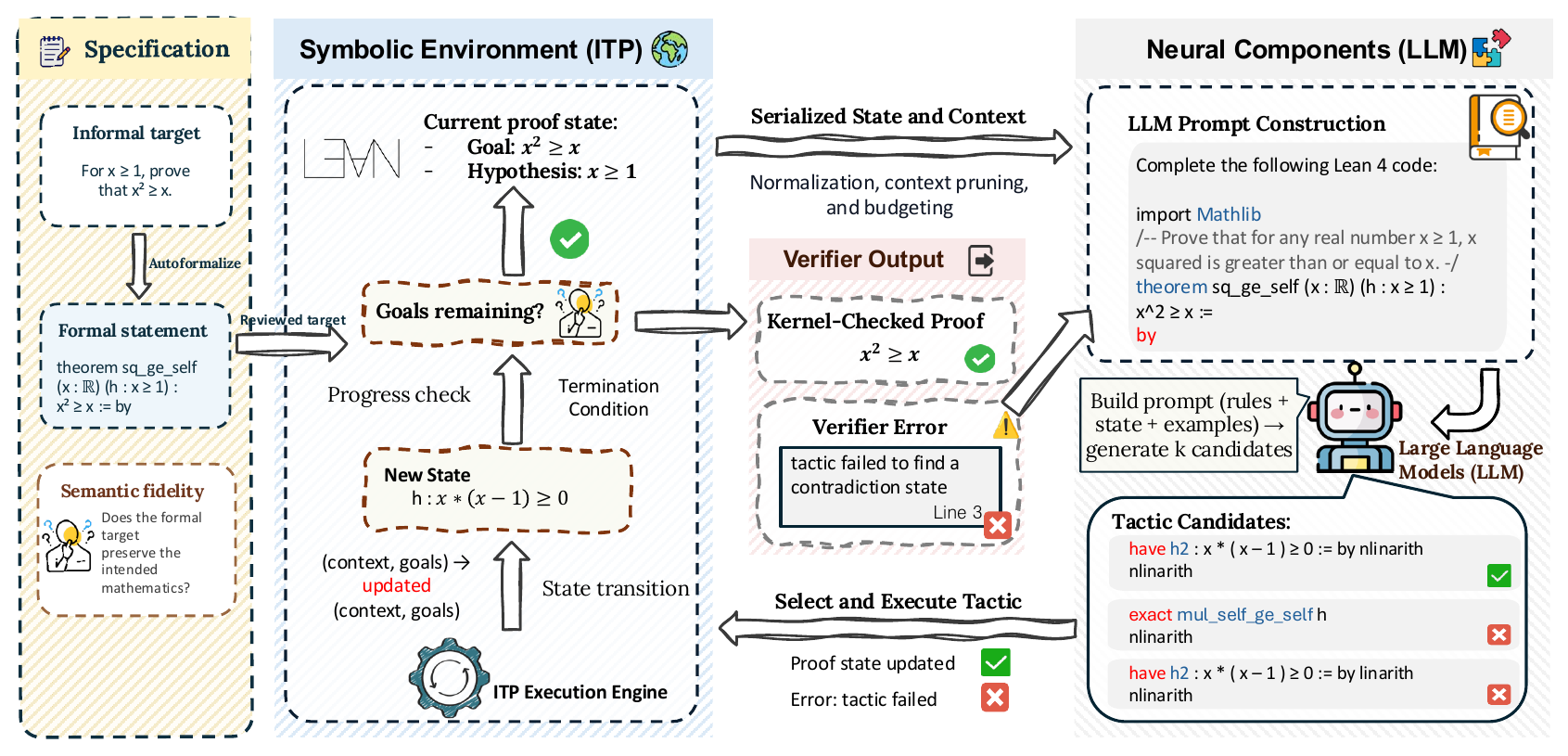}
    \caption{\textbf{Neural and symbolic components in an interactive
    theorem-proving workflow.} An informal target is reviewed during
    specification and translated into a formal statement. The proof assistant
    maintains the formal state and checks candidate proofs, while the language
    model receives a serialized context and proposes tactics. Semantic review
    asks whether the formal statement preserves the intended mathematics.
    Kernel verification instead determines whether the proof is accepted in
    the stated environment.}
    \label{fig:provisional-neural-symbolic-loop}
\end{figure}

The work supported by these systems now spans several scales. Benchmarks such as miniF2F and PutnamBench evaluate proof generation for specified statements; repository-based tasks add dependencies, project conventions, and requirements that extend beyond closing an individual goal~\citep{zheng2022minif2f,tsoukalas2024putnam,letson2026sorrydb,xie2026mathlibpr}. At project scale, formalization requires coordinating definitions and intermediate results across files while preserving the correspondence between a mathematical source and its formal development~\citep{zhang2026leanmarathon}. Interactive workspaces and shared research platforms support this process by exposing intermediate artifacts, retaining context, and enabling human intervention~\citep{LeaTeam2026Software,moakhar2026theo,chen2026prove2me}. Together, these developments make sustained coordination across model, tool, and human interactions part of the mathematical task itself.

Each setting calls for a different interpretation of success. A kernel-accepted proof establishes derivability of a particular formal statement relative to the declared logic, axioms, and dependencies. Assessing its mathematical contribution also requires asking whether the statement faithfully encodes the intended problem, whether the result is new, and how the proof was obtained. Autoformalization makes the first question especially important: a compilable and provable translation can still alter the mathematical meaning of its source~\citep{CornishEtAl2026FaithformBenchArXiv260810916}. Research claims place additional demands on prior-art search, attribution, exposition, and expert judgment. The reported study of AI-assisted work on Erd\H{o}s problems, for example, distinguishes contributions involving existing literature from other forms of progress, while First Proof evaluates correctness, exposition, and attribution separately~\citep{feng2026semi,abouzaid2026thirdbatch}. Proof checking, artifact inspection, and mathematical review thus answer complementary questions about a result.

We argue that \emph{the versioned workflow is the appropriate unit of analysis for research-scale AI mathematics}. Evaluation should account for the target statement, model and tool configuration, proof environment, intermediate artifacts, resource use, human interventions, and review history. These records make it possible to distinguish outcomes that a final answer or success rate alone can obscure. Rebuilding a formal artifact, rerunning the process that produced it, and independently reconstructing its mathematical argument are distinct forms of reproducibility. Similarly, benchmark accuracy, repository acceptance, project completion, and an open research claim describe different achievements, rather than successive levels on a single capability scale. Explicitly identifying the evaluation object and the evidence supporting it provides a basis for comparison while keeping conclusions tied to what was demonstrated.

We develop this argument through a narrative, non-exhaustive survey of literature and public records. The survey centers on LLM-based formal mathematics: data construction, autoformalization, proof synthesis, and the workflows that connect these components. Informal reasoning systems and domain-specific solvers serve as comparison points; selected research records and community platforms illustrate the additional requirements of sustained mathematical work. Complementing existing surveys of mathematical reasoning and autoformalization~\citep{weng2025autoformalization,JuDong2026AI4MathSurvey}, we connect technical methods to the evidence needed to assess their outputs. The dated cases ground this analysis without providing a field-wide estimate of research success or a validated ranking of systems.

The paper makes three contributions:
\begin{enumerate}[leftmargin=1.5em, itemsep=3pt, topsep=4pt, parsep=0pt, partopsep=0pt]
    \item \textbf{A taxonomy of methods and research workflows.} We review formal-mathematics tasks and data, organizing methods along three overlapping axes: training strategies, inference-time search and adaptation, and agentic workflows. We relate these methods to benchmark protocols, project-scale developments, and collaborative infrastructure.
    \item \textbf{An evidence framework for heterogeneous mathematical claims.} We distinguish four evaluation objects: benchmark theorems, repository tasks, formalization projects, and open research claims. We organize the evidence supporting these claims along six dimensions: target and specification fidelity; correctness and verification scope; workflow and resource disclosure; artifact availability and reproducibility; human contribution and provenance; and originality, significance, and external review. A reporting checklist translates these dimensions into concrete disclosure requirements while preserving their distinct roles.
    \item \textbf{A research agenda grounded in the evidence framework.} We identify priorities in data and evaluation, relational representations, conjecture and concept generation, external-tool integration, and human--AI collaboration. For each, we discuss what evidence would support progress beyond isolated proof success.
\end{enumerate}

Section~\ref{sec:foundations} introduces the foundations, tasks, and data, and Section~\ref{sec:methodologies} reviews the methodological axes. Sections~\ref{sec:sota_benchmarks} and~\ref{sec:sep2026-update} examine benchmark results and selected research records. Section~\ref{sec:community-platforms} discusses platforms and community infrastructure. Section~\ref{sec:literature-map-2026} develops the evaluation framework and reporting checklist, followed by research challenges in Section~\ref{sec:challenges} and conclusions in Section~\ref{sec:conclusion}.

\section{Foundations and Preliminaries}
\label{sec:foundations}

\revadd{This section briefly sketches automated theorem proving, introduces
proof assistants and model-containing systems for mathematics, and defines the
core tasks, workflows, datasets, and evaluation conventions used throughout the
survey.}


\subsection{Historical Context of Automated Theorem Proving}

\begin{revisionblock}
Projects to systematize calculation and reasoning long predate electronic computing. Leibniz's
1666 \emph{Dissertatio de arte combinatoria} contains an early sketch of a
universal characteristic and logical calculus; his related projects for a
\emph{characteristica universalis} and \emph{calculus ratiocinator} remained
programmatic rather than a working proof checker
~\citep{Leibniz1666DeArteCombinatoria,Peckhaus2024LeibnizLogicInfluence}.
These proposals are therefore best read as intellectual precursors, not as implementations
of modern formal verification.  Nineteenth- and twentieth-century formal logic
subsequently supplied precise calculi through works such as Frege's
\emph{Begriffsschrift} and Whitehead and Russell's \emph{Principia Mathematica}
~\citep{Frege1879Begriffsschrift,WhiteheadRussell1910Principia}.
Gödel's incompleteness theorems then showed a specific limitation: any
consistent, effectively axiomatized formal theory strong enough to express basic
arithmetic is incomplete; they do not imply that every formal system is equally
limited or that proof checking is impossible~\citep{Goedel1931Unentscheidbare}.
\end{revisionblock}

\revadd{Newell and Simon's 1956 report on the Logic Theory Machine described
heuristic search for proofs in symbolic logic~\citep{newell1956logic}.
Its symbolic representation and search procedures provided an early concrete
demonstration of automated proof discovery.}

\revadd{In 1965, John Alan Robinson introduced the resolution principle.
Resolution is refutation-complete for first-order logic: under a fair search,
an unsatisfiable clause set admits a derivation of contradiction. Equivalently,
it yields a semidecision procedure for validity, not a decision procedure for
arbitrary first-order formulas---search on a non-valid formula need not
terminate~\citep{robinson1965resolution}. Its use of clausal normal form and a
uniform inference rule established a foundation for later automated theorem
provers and remains influential in modern systems.}

\revadd{Subsequent work produced several paradigms for automated theorem proving (ATP), each addressing different parts of the theorem-proving task. Saturation-based provers such as E~\citep{schulz2019e}, Vampire~\citep{kovacs2013vampire}, and SPASS~\citep{weidenbach2001spass} systematically derive consequences from axioms using sophisticated term orderings and redundancy elimination techniques. These provers have been widely used in automated reasoning and as backends for other proof systems.}

\revadd{A parallel line of work pursued decision procedures for restricted
languages. SAT solvers decide propositional satisfiability. Satisfiability
modulo theories (SMT) solvers such as Z3 and cvc5 combine Boolean search with
theory procedures for domains such as arithmetic, arrays, and bit-vectors
~\citep{demoura2008z3,barbosa2022cvc5}. Their automation is powerful within
the supported fragments, but it does not turn unrestricted first-order or
higher-order theorem proving into a decision problem.}

\begin{revisionblock}
Interactive theorem provers (ITPs), or proof assistants, developed as a
complementary paradigm in which users and automation construct formal artifacts
for machine checking.  Representative systems include Mizar~\citep{trybulec1985mizar},
HOL-family provers~\citep{gordon1993hol}, Isabelle~\citep{paulson1994isabelle},
Lean~\citep{moura2021lean4}, and the Rocq Prover (formerly the Coq Proof
Assistant)~\citep{barras1999coq,RocqDevelopmentTeam2025Renaming}.  We use the
    current name \emph{Rocq} below, retaining \emph{Coq} only in historical
    publication titles and benchmark or project names. A kernel or verifier checks the
    exact artifact relative to the system's logic, axioms, trusted implementation,
    and pinned library configuration. Such checking establishes acceptance of
    that exact artifact in the disclosed environment; it does not by itself establish
that the encoding matches the intended theorem, is novel, or is mathematically
significant.  Formalizations of the Four Color Theorem and Odd Order Theorem in
Rocq~\citep{gonthier2008formal,gonthier2013odd}, and of the Kepler Conjecture in
HOL Light and Isabelle~\citep{hales2017formal}, illustrate the reach of this
approach while also documenting substantial expert effort in specification,
library construction, proof engineering, and maintenance.
\end{revisionblock}

\subsection{Formal Mathematics and ITPs}
\begin{revisionblock}
We focus on ITP-based formal mathematics, in which encoded propositions and
derivations are machine-checked relative to a specified logic, trusted
implementation, and library snapshot. In
tactic-oriented interfaces, a \textbf{proof state} records goals and local
hypotheses, and tactics such as \texttt{intro}, \texttt{apply}, \texttt{rewrite},
and \texttt{induction} propose state transformations.  Tactics may be untrusted
metaprograms: acceptance ultimately depends on the kernel checking the generated
proof term or theorem.  Libraries such as Lean's mathlib and Rocq's Mathematical
Components provide definitions and lemmas above this foundational core
~\citep{mathlib2020,vandoorn2020mathlib,mahboubi22mathcomp}.

No single taxonomy captures modern proof systems: logical foundation, mode of
interaction, and degree of automation are distinct axes.  The following is a
selective map of general-purpose assistants and proof frameworks by logical
foundation:

\begin{itemize}[leftmargin=1.3em, topsep=2pt]
    \item[\ding{182}] \textbf{Dependent type theories.}  Lean implements a
    dependent type theory, while Rocq is based on the Calculus of Inductive
    Constructions~\citep{demoura2015lean,moura2021lean4,barras1999coq,bertot2013interactive}.
    \item[\ding{183}] \textbf{Higher-order logic.}  HOL-family provers and
    Isabelle/HOL use variants of classical higher-order logic.  Isabelle also
    supports other object logics, so \emph{Isabelle} and \emph{Isabelle/HOL}
    should not be treated as synonyms~\citep{gordon1993hol,paulson1994isabelle}.

    \item[\ding{184}] \textbf{First-order computational logic.}  ACL2 combines
    a first-order logic of recursive functions with an automated prover and an
    interactive development environment~\citep{kaufmann1996acl2}; this differs
    architecturally from LCF-style proof assistants even though both support
    machine-checked developments.

    \item[\ding{185}] \textbf{Predicate subtyping.}  PVS uses higher-order logic
    enriched with dependent and predicate subtypes~\citep{owre1992pvs}.  This is
    a description of its specification language, not a point on a scale from
    ``HOL'' to ``dependent type theory.''

    \item[\ding{186}] \textbf{Minimal proof frameworks.}  Metamath specifies and
    checks user-supplied formal systems; its main \texttt{set.mm} database develops
    classical first-order logic and Zermelo--Fraenkel set theory with the axiom of choice (ZFC)~\citep{metamath}.  Its explicit
    proof format made it an early target for neural proof generation
    ~\citep{whalen2016holophrasm,wang2020learning,polu2020generative}.
\end{itemize}

Automation is a separate design axis from logical foundation.  E and Vampire are standalone
first-order ATPs, whereas Isabelle's Sledgehammer selects facts, invokes external
provers, and attempts to reconstruct a proof in Isabelle
~\citep{schulz2002brainiac,kovacs2013vampire,paulson2012sledgehammer}.  Internal
tactics and decision procedures form yet another integration pattern.  Thus
labels such as ``high'' or ``medium'' automation are not stable system properties:
they depend on the theory, libraries, tactics, external tools, and resource budget.

Domain-specific symbolic solvers form a separate category from general-purpose
ITPs.  Geometry provers based on deductive databases, algebraic reasoning, or
Wu-style methods operate over restricted geometric languages
~\citep{chou1993automated,chou2000deductive,wu2008decision}.  AlphaGeometry is a
neuro-symbolic instance: a transformer proposes auxiliary constructions in a
symbolic geometry language, and an engine combining a deductive database with
algebraic reasoning derives and checks consequences~\citep{trinh2024alphageometry}.  It does
not take a diagram image as its problem input and should not be classified as a
multimodal model or a general-purpose ITP.
\end{revisionblock}

\subsection{Foundation Models for Mathematical Reasoning}

\revadd{Foundation models do not define a single proving architecture. A
model may generate prose or code, propose tactics, retrieve premises, score
states, or translate specifications, and only some pipelines submit a formal
artifact to a checker. We therefore use four overlapping categories---general-purpose,
mathematics-specialized, reasoning-oriented, and formal-artifact
pipelines---of model-containing systems based on training and
operational role rather than presumed capability.}

\begin{revisionblock}
Public evaluations of general-purpose LLMs such as GPT-4, Gemini, Llama, and
Qwen report results on selected mathematical tasks under the authors' own
protocols; those results depend on the benchmark, prompt, tools, and inference
budget~\citep{achiam2023gpt4,team2023gemini,reid2024gemini15,
touvron2023llama2,dubey2024llama3,yang2024qwen}. Chain-of-thought and zero-shot
step-by-step prompting provide early evidence that elicitation format can
materially change benchmark performance~\citep{wei2022chain,kojima2022zero}.
These observations motivate mathematical specialization and tool coupling, but
the cited reports do not establish reliability beyond their stated tasks and
protocols.
\end{revisionblock}

\begin{revisionblock}
Mathematics-specialized models undergo continued pretraining on mathematical
corpora. LLEMMA reports gains from a corpus assembled from mathematical papers,
web material, and code, while DeepSeekMath reports benchmark gains after
large-scale pretraining on mathematical web data
~\citep{azerbayev2024llemma,shao2024deepseekmath}. These source-reported gains
motivate specialized pretraining in downstream theorem-proving experiments but
remain conditional on the authors' tasks and protocols.
\end{revisionblock}

\revadd{Reports for reasoning-oriented models such as o1, o3-mini, and
DeepSeek-R1 describe additional inference-time reasoning and mathematics
evaluations under model-specific settings
~\citep{jaech2024openai,openai2024o3,deepseekr1}. The models differ in
disclosed internals, tool access, sampling, reasoning effort, and contamination
controls. Their reported scores therefore motivate targeted tests of search and
verifier feedback in formal environments, but they are not a common measure of
expert-level mathematical ability.}

\begin{revisionblock}
Formal-artifact pipelines form a fourth overlapping category,
but they do not all target the same kind of proof environment. GPT-f combined a transformer policy with
proof search for the Metamath language; generated steps were accepted only when
the Metamath verifier checked the completed proof.  Its authors report that 23
shorter proofs for existing theorems were accepted into the main Metamath library
~\citep{polu2020generative}.  This was a concrete library contribution, but it
should not be paraphrased as discovering new theorem statements or as operating
in a tactic-based ITP.  The DeepSeek-Prover series, Goedel-Prover, and
Seed-Prover instead report results in Lean under their respective data, search,
and evaluation protocols
~\citep{xin2024deepseekprover,xin2024deepseekproverv15,ren2025deepseekproverv2,lin2025goedel,lin2025goedelv2,chen2025seed}.

Domain-specific neuro-symbolic solvers should be reported separately.  In the
IMO 2024 evaluation, AlphaProof handled the non-geometry problems in Lean and
AlphaGeometry~2 handled geometry in its specialized symbolic setting; together
they received 28 of 42 points, a silver-medal-equivalent total, after experts
translated the problems into the systems' formal representations
~\citep{hubert_olympiad-level_2025,GoogleDeepMind2024IMOSilverAnnouncement}.  Results across these systems are not directly
interchangeable: the target language, statement-translation protocol, library,
inference budget, and evaluation unit all differ.
\end{revisionblock}

\Needspace{7\baselineskip}
\subsection{Informal Mathematical Reasoning}

\revadd{Although this survey's formal-methods discussion centers on theorem
proving with machine-checked proof artifacts, informal reasoning offers
methodological precedents for explicit
intermediate reasoning and aggregation~\citep{wei2022chain,wang2023selfconsistency},
tree- and graph-structured search~\citep{yao2023tree,besta2024graph}, iterative
refinement~\citep{madaan2023selfrefine}, and program-aided tool use
~\citep{gao2023pal}. Transfer must nevertheless be evaluated
inside the target proof environment: scores or natural-language critiques
produced by learned models do not have the same evidentiary status as a proof
assistant's check of an elaborated formal artifact relative to an explicit
trusted base.}

\begin{revisionblock}
Informal and formal benchmarks therefore test different objects
~\citep{hendrycks2021measuring,he2024olympiadbench,glazer2024frontiermath,
zheng2022minif2f,tsoukalas2024putnam}.  Exact-answer or rubric-based gains do
not by themselves demonstrate formal proving ability, and comparisons across model generations must
account for changes in training data, prompting, tools, contamination controls,
and inference budget. Mathematics-specialized models may still guide a formal
pipeline~\citep{shao2024deepseekmath}. The resulting proof claim, however,
should be assessed through a faithful target and replayable checked artifact
~\citep{hubert_olympiad-level_2025,ren2025deepseekproverv2}.
\end{revisionblock}


\subsection{Autoformalization}
\label{sec:autoformalization_main}
\begin{revisionblock}
Autoformalization bridges informal mathematical intent and a machine-checkable
target; it is therefore a specification task, not merely a change of notation.
A useful output must be accepted by the target environment and preserve the
intended mathematics. Early
LLM autoformalization work studied natural-language-to-Isabelle translation
and the use of translated statements as prover-training data
~\citep{wu2022autoformalization}. Thor illustrates a related but distinct task:
it couples a language model to external automated provers for theorem proving
~\citep{jiang2022thor}. Later
work explores multi-agent decomposition~\citep{zhang2025masa}, thought-guided
example retrieval for logical autoformalization~\citep{hu2025ltrag}, iterative
semantic repair~\citep{chen2025reform}, and process-based use of compiler
feedback~\citep{lu2024process}; \citet{weng2025autoformalization} provide a
broader survey.
These releases address different data and evaluation bottlenecks. ATLAS builds
paired informal--formal statement data for translation, whereas QDTSynth
synthesizes and filters Lean~4 theorem statements for prover training
~\citep{liu2025atlas,wang2025qdtsynth}. Herald adds natural-language
annotations to Mathlib4 and trains a translator; the Lean IMO dataset releases
complete proofs and decomposed lemmas for a bounded set of IMO problems
~\citep{gao2024herald,yousefzadeh2025leandataset}. FormalAlign is instead a
learned evaluator of informal--formal alignment; its score may reduce review
cost but does not itself establish semantic equivalence
~\citep{lu2024formalalign}.
\end{revisionblock}

\begin{revisionblock}
A formal statement can elaborate successfully while still mistranslating its
informal source. Recent work therefore intervenes at different points in the
specification pipeline: conceptual or dependency retrieval
~\citep{lu2026cramf,wang2026ddr}, decomposition and proof-flow modeling
~\citep{min2026dna,cabral2026proofflow}, diagnostic feedback
~\citep{peng2026criticlean,wang2026formalrx}, and theory-level construction
~\citep{min2026theorylevel}. Because these systems use different tasks and
evaluators, Section~\ref{sec:sep2026-update} compares their evaluation objects
rather than treating syntactic-acceptance and semantic-alignment scores as
interchangeable.
\end{revisionblock}

\subsection{\revadd{Core Tasks and the Prover Interaction Loop}}
\label{sec:core_tasks}

\begin{revisionblock}
Many neural provers that operate inside an ITP use an iterative prover
interaction loop: a model proposes one or more actions, the environment executes
them, and the resulting successor state or diagnostic informs the next
decision. A result is accepted only when the environment's trusted checker
validates the resulting artifact.
Figure~\ref{fig:provisional-neural-symbolic-loop} separates semantic review
from kernel validation and identifies where model-generated tactics enter the
checked interaction loop.
Autoformalization is an upstream task rather than a necessary stage of every
prover---many benchmarks supply an already formal statement. We therefore
distinguish two related tasks, \textbf{Autoformalization} and \textbf{Proof
Generation}, without assuming that every system performs both.
\end{revisionblock}


\begin{revisionblock}
Given a formal statement, \textbf{Proof Generation} searches for a proof term or
script that the target environment accepts.  Some systems explicitly separate
premise selection, tactic generation, and search~\citep{wu2021tacticzero,jiang2022thor};
others generate larger proof fragments or whole scripts, sometimes interleaved
with decomposition and verifier feedback
~\citep{ren2025deepseekproverv2,chen2025seed,lin2025goedelv2}.  These are design
choices rather than a simple chronological replacement of modular methods by a
single end-to-end paradigm. The next two subsections make the target
environments and evaluation objects explicit; Section~\ref{sec:methodologies}
then develops the methodological taxonomy.
\end{revisionblock}

\vspace{0.8em}
\subsection{\revadd{Comparison of Formal Proof Environments}}

\begin{revisionblock}
The target environment and exposed interaction layer jointly determine the
object language, trusted base, available library, interaction protocol, and
action space of a neural prover.
Table~\ref{tab:itp_comparison} compares these design-relevant axes for a selected
set of systems.  It is neither an exhaustive catalog nor a capability ranking.
\end{revisionblock}

\begin{revisionblock}
\begin{table}[h]
\ifshowrevisions\color{revisionblue}\fi
\renewcommand{\arraystretch}{1.05}
\centering
\begingroup
\fontsize{8}{9.2}\selectfont
\setlength{\tabcolsep}{2.5pt}
\begin{tabularx}{\textwidth}{@{}>{\raggedright\arraybackslash}p{0.11\textwidth}
>{\raggedright\arraybackslash}p{0.15\textwidth}
>{\raggedright\arraybackslash}p{0.19\textwidth}
>{\raggedright\arraybackslash}p{0.205\textwidth}X@{}}
\toprule
\rowcolor{gray!15} \textbf{System} & \textbf{Logical basis} &
\makecell[l]{\textbf{Checked artifact}\\\textbf{and interface}} &
\textbf{Automation pattern} &
\makecell[l]{\textbf{Survey-relevant library}\\\textbf{or machine-learning resource}} \\
\midrule
Isabelle/HOL & HOL as an object logic in the generic Isabelle framework &
Isar proofs and tactic commands produce kernel-checked theorems &
Internal proof methods; Sledgehammer can select facts and call external ATPs &
AFP; LISA~\citep{jiang2021lisa}; IsarStep~\citep{li2020isarstep} \\
Rocq (formerly Coq) & Calculus of Inductive Constructions (CIC) &
Gallina terms and tactic scripts elaborated to kernel-checked proof terms &
Tactics and decision procedures; plugins may emit kernel-checkable terms or enlarge the trusted code base &
Mathematical Components~\citep{mahboubi22mathcomp}; CoqGym~\citep{yang2019coqgym} \\
HOL Light & Classical higher-order logic &
OCaml tactics construct theorems through a small LCF-style kernel &
Tactic automation and integrations with external first-order reasoning &
HOList~\citep{bansal2019holist,paliwal2019graph} \\
Metamath & User-declared systems; \texttt{set.mm} develops classical first-order logic (FOL) and ZFC &
Explicit substitution-based proof sequences checked by a small verifier &
Search is external to the verifier &
Holophrasm~\citep{whalen2016holophrasm}; GPT-f~\citep{polu2020generative} \\
Lean 4 & Dependent type theory &
Terms and tactic scripts elaborated by the frontend to kernel-checked proof terms &
Built-in and library tactics plus external search and agent tooling &
mathlib4~\citep{Mathlib2026Lean4Repository}; LeanDojo~\citep{yang2023leandojo};
\mbox{LeanDojo-v2}~\citep{hsiang2025leandojov2} \\
\bottomrule
\end{tabularx}
\endgroup
\caption{\revadd{Comparison of selected proof environments used in learning-based
theorem proving. The columns separate logical foundation, checked artifact, and
automation architecture. Qualitative ``high/medium/low'' automation labels and
raw library-size counts are omitted because such labels and counts are not
comparable without a fixed task, release, configuration, and counting rule.
Historical resource names such as CoqGym are retained.}}
\label{tab:itp_comparison}
\end{table}
\end{revisionblock}

\begin{revisionblock}
Table~\ref{tab:itp_comparison} should be read as comparing system stacks, not
only proof languages. In Lean, model behavior depends on mathlib and the
interaction layer exposed by tools such as LeanDojo
~\citep{mathlib2020,yang2023leandojo}. LeanDojo-v2 extends this infrastructure
into an end-to-end Lean 4 toolkit for repository tracing, dataset management,
model training, and interactive proof search~\citep{hsiang2025leandojov2}. In
Isabelle, Sledgehammer may call
external ATPs and reconstruct a proof~\citep{paulson2012sledgehammer}; GPT-f
instead searches Metamath's explicit proof language
~\citep{polu2020generative}. Cross-system solve rates consequently entangle
model behavior with library coverage, available automation, interface design,
and resource budget, and should not be interpreted as model-only comparisons.
\end{revisionblock}

\subsection{Datasets and Evaluation}
\label{sec:datasets_benchmarks}

\begin{revisionblock}
Once the target environment is fixed, evaluation still depends on the corpus or
benchmark that defines the task and on how its units are counted. Corpus-scale
comparisons additionally require a common tokenization, an explicit release
snapshot, a deduplication policy, and a clear distinction among statements,
proofs, trajectories, and tokens. We therefore do not collapse web corpora,
library exports, and synthetic theorem--proof data into a single scale table.
The cited releases illustrate why raw corpus sizes do not share a denominator:
Lean-GitHub extracts formal data from Lean repositories; LeanDojo traces proof
states and actions from mathlib; Lean Workbook and FormalMATH publish
problem- or statement-level resources; and DeepSeek-Prover constructs
synthetic formal statements with generated proofs
~\citep{wu2024lean,yang2023leandojo,ying2024leanworkbook,
yu2025formalmath,xin2024deepseekprover}. A reproducible comparison must name
the release and state whether its count denotes repositories, declarations,
statements, proofs, theorem--proof pairs, or interaction trajectories.

\end{revisionblock}

\begin{revisionblock}
\begin{table}[h]
\ifshowrevisions\color{revisionblue}\fi
\renewcommand{\arraystretch}{1.03}
\centering
\begingroup
\fontsize{7.6}{8.6}\selectfont
\setlength{\tabcolsep}{2.2pt}
\begin{tabularx}{\textwidth}{@{}>{\raggedright\arraybackslash}p{0.175\textwidth}
>{\centering\arraybackslash}p{0.055\textwidth}
>{\raggedright\arraybackslash}p{0.15\textwidth}
>{\raggedright\arraybackslash}p{0.125\textwidth}
>{\raggedright\arraybackslash}p{0.17\textwidth}X@{}}
\toprule
\rowcolor{gray!15} \textbf{Resource} & \textbf{Release} &
\makecell[l]{\textbf{Evaluation}\\\textbf{object}} &
\makecell[l]{\textbf{Formal}\\\textbf{environment(s)}} &
\makecell[l]{\textbf{Stated size}\\\textbf{and counting unit}} & \textbf{Scope} \\
\midrule
\textbf{miniF2F}~\citep{zheng2022minif2f} & 2022 & Formal-statement proof search &
Metamath and Lean; partial Isabelle and HOL Light &
488 statements (244 validation + 244 test) & Competition and course-level problems \\
\textbf{ProofNet}~\citep{azad2023proofnet} & 2023 & Statement autoformalization and proof &
Lean 3 & 371 aligned examples & Each example pairs an informal statement and proof with a formal statement \\
\textbf{PutnamBench}~\citep{tsoukalas2024putnam} & 2024 & Formal-statement proof search &
Lean 4, Isabelle, Rocq (named Coq in the cited release) & 640 theorems; 1,692 language-specific formalizations & Undergraduate competition problems \\
\textbf{ProverBench}~\citep{ren2025deepseekproverv2} & 2025 & Formal-statement proof search &
Lean 4 & 325 formalized problems & Textbook problems; 15 selected AIME 2024--2025 items \\
\textbf{CombiBench}~\citep{liu2025combibench} & 2025 & Proof and fill-in-the-blank evaluation &
Lean 4 & 100 problems & Combinatorics across school, IMO, and university levels \\
\textbf{FormalMATH}~\citep{yu2025formalmath} & 2025 & Formal-statement proof search &
Lean 4 & 5,560 Lean statements & High-school Olympiad through undergraduate mathematics \\
\textbf{Ineq-Comp}~\citep{zhao2025ineqcomp} & 2025 & Robustness under controlled composition &
Lean 4 & 275 fixed examples & 75 seeds, 150 controlled variants, and 50 collected inequality problems \\
\textbf{FATE-H/X}~\citep{jiang2025fate} & 2025 & Formal-statement proof search &
Lean 4 & 200 problems (100 H + 100 X) & Advanced abstract and commutative algebra \\
\textbf{Formal Conjectures}~\citep{FormalConjecturesAuthors2026Repository} & 2025; living &
Research-statement repository & Lean 4 & Living collection; item count changes with repository state & Formalized research conjectures and open problems \\
\textbf{FormalProofBench}~\citep{ravi2026formalproofbench} & 2026 & Agentic formal-statement proof search &
Lean 4 & 200 problems & Advanced undergraduate and graduate mathematics \\
\textbf{CAM-Bench}~\citep{long2026cambench} & 2026 & Formal-statement proof search with recovered local context &
Lean 4 & 1,000 proof targets & Optimization, numerical linear algebra, and numerical analysis \\
\textbf{MA-ProofBench}~\citep{pu2026maproofbench} & 2026 & Formal-statement proof search &
Lean 4 & 200 theorems (100 Level I + 100 Level II) & Mathematical analysis from undergraduate to PhD-qualifying level \\
\bottomrule
\end{tabularx}
\endgroup
\caption{\revadd{Selected resources with machine-checkable formal targets. Counts
use the units reported by each cited release; theorems, language-specific
formalizations, aligned informal--formal examples, and living repository items
are not interchangeable units. This table is not a leaderboard; system results
also depend on the exact split, library snapshot, permitted tools, inference
budget, and contamination controls. Ineq-Mix can generate additional composed
inequalities beyond the fixed Ineq-Comp set; reproducible uses of Formal
Conjectures should identify an immutable benchmark tag or repository commit.}}
\label{tab:benchmark_comparison}
\end{table}
\end{revisionblock}

\begin{revisionblock}
Table~\ref{tab:benchmark_comparison} makes the evaluation object and counting
unit explicit; these resources are not directly rankable.  In particular,
Formal Conjectures is a changing repository rather than a fixed hidden test set,
and informal resources such as FrontierMath answer a different evaluation
question~\citep{FormalConjecturesAuthors2026Repository,
glazer2024frontiermath}. Recent benchmarks also widen the mathematical and
operational scope. FormalProofBench uses a private set of advanced undergraduate
and graduate problems with an agentic Lean harness; CAM-Bench tests
context-dependent problems in computational and applied mathematics; and
MA-ProofBench separates undergraduate from PhD-qualifying analysis
~\citep{ravi2026formalproofbench,long2026cambench,pu2026maproofbench}.
Lean-IMO-Bench instead targets short statements whose proofs require
non-routine, multi-step Olympiad reasoning~\citep{kung2026leap}. These resources
do not define a common leaderboard. Comparisons should name the dataset
revision, formal-library snapshot, tool access, interaction limit, metric, and
inference budget.
\end{revisionblock}

\revadd{With target artifacts, interaction loops, and evaluation units now
fixed, we next compare how learning-based methods generate actions, allocate
search, and use verifier feedback.}

\section{\revadd{Methods for LLM-Based Formal Theorem Proving}}
\label{sec:methodologies}


\revadd{We organize LLM-based formal theorem proving along three overlapping
axes. Training strategies determine how a policy or critic is fitted;
inference-time mechanisms determine how candidates are generated, ranked, and
searched; and agentic workflows determine how state, tools, and feedback are
coordinated across calls. A single system may instantiate all three, so the
categories are not mutually exclusive. Geometry-specific systems appear only
as contrast cases because their representations and symbolic engines define a
different task regime. Figure~\ref{fig:ntp-taxonomy-outline} summarizes these
axes and illustrates why individual systems may occupy more than one branch.
Section~\ref{sec:challenges} synthesizes limitations that cut across these
axes.}

\definecolor{outlineDraw}{RGB}{120,120,120}
\definecolor{ntpBlue}{RGB}{216,226,234}       
\definecolor{ntpRed}{RGB}{247,226,231}        
\definecolor{ntpYellow}{RGB}{252,238,221}     
\definecolor{ntpGreen}{RGB}{204,231,207}      

\newcommand{\citeleaf}[1]{{\normalsize #1}}

\forestset{
  outline-box/.style={
    rectangle,
    draw=outlineDraw,
    line width=0.7pt,
    rounded corners=3pt,
    inner xsep=5pt,
    inner ysep=4pt,
    minimum height=1.6em,
  },
  outline-leaf/.style={
    outline-box,
    text width=36em,
    align=justify,
    font=\normalsize,
  },
  ver/.style={
    rotate=90,
    child anchor=north,
    parent anchor=south,
    anchor=center,
    transform shape,
  },
}

\begin{figure*}[!ht]
\centering
\resizebox{\textwidth}{!}{%
\begin{forest}
forked edges,
for tree={
  grow=east,
  reversed=true,
  anchor=base west,
  parent anchor=east,
  child anchor=west,
  base=center,
  font=\large,
  outline-box,
  edge={draw=outlineDraw, line width=0.7pt},
  s sep=12pt,
  l sep=10pt,
  inner xsep=5pt,
  inner ysep=4pt,
},
where level=1{
  text width=18.5em,
  minimum height=2.6em,
  font=\large,
  align=center
}{},
where level=2{
  text width=18.5em,
  minimum height=2.6em,
  font=\large,
  align=center
}{},
where level=3{
  text width=36em,
  font=\normalsize,
  calign with current edge,
  edge path={
    \noexpand\path[\forestoption{edge}]
      (!u.parent anchor) -- (.child anchor)\forestoption{edge label};
  }
}{},
[
  {\shortstack{\textbf{Neural Theorem Proving}\\\textbf{with LLMs}}},
  font=\large,
  ver,
  fill=ntpBlue,
  xshift=10mm,
  s sep=22pt
  [
    {\textbf{Training Strategies}},
    fill=ntpRed
    [
      {\textbf{Supervised Fine-Tuning}},
      fill=ntpRed!70
      [
        {\citeleaf{%
          DeepSeek-Prover~\citep{xin2024deepseekprover},
          Goedel-Prover~\citep{lin2025goedel},
          TheoremLlama~\citep{wang2024theoremllama},
          Leanabell~\citep{zhang2025leanabell},
          ALCHEMY~\citep{wu2025alchemy}%
        }},
        fill=ntpRed!15
      ]
    ]
    [
      {\textbf{Reinforcement Learning}},
      fill=ntpRed!70
      [
        {\citeleaf{%
          DeepSeek-Prover-V1.5~\citep{xin2024deepseekproverv15},
          Goedel-Prover-V2~\citep{lin2025goedelv2},
          Kimina-Prover~\citep{wang2025kimina},
          Leanabell-V2~\citep{ji2025leanabellv2},
          Seed-Prover~\citep{chen2025seed},
          DeepSeek-Prover-V2~\citep{ren2025deepseekproverv2},
          Seed-Prover~1.5~\citep{chen2025seedprover15masteringundergraduatelevel},
          Abel~\citep{gloeckle2024abel},
          AlphaProof~\citep{hubert_olympiad-level_2025},
          Process-Verified RL~\citep{kim2026processverified}%
        }},
        fill=ntpRed!15
      ]
    ]
    [
      {\textbf{Search-in-the-loop Training}},
      fill=ntpRed!70
      [
        {\citeleaf{%
          DeepSeek-Prover~\citep{xin2024deepseekprover},
          HunyuanProver~\citep{li2024hunyuan},
          HTPS~\citep{lample2022hypertree},
          InternLM2.5-StepProver~\citep{wu2024internlm}%
        }},
        fill=ntpRed!15
      ]
    ]
  ]
  [
    {\textbf{Inference-Time Mechanisms}},
    fill=ntpYellow
    [
      {\textbf{Planning \& Decomposition}},
      fill=ntpYellow!70
      [
        {\citeleaf{%
          DSP~\citep{jiang2023draftsketch,cao2025reviving},
          DeepSeek-Prover-V2~\citep{ren2025deepseekproverv2},
          SubgoalXL~\citep{zhao2024subgoalxl},
          Seed-Prover~1.5~\citep{chen2025seedprover15masteringundergraduatelevel},
          LEAP~\citep{kung2026leap},
          Goedel-Architect~\citep{chung2026goedelarchitect}%
        }},
        fill=ntpYellow!15
      ]
    ]
    [
      {\textbf{Premise and Example Retrieval}},
      fill=ntpYellow!70
      [
        {\citeleaf{%
          ReProver~\citep{yang2023leandojo},
          COPRA~\citep{thakur2024context},
          Hilbert~\citep{varambally2025hilbert},
          LemmaHead~\citep{yang2025lemmahead},
          LTRAG~\citep{hu2025ltrag}%
        }},
        fill=ntpYellow!15
      ]
    ]
    [
      {\textbf{Ranking and Search}},
      fill=ntpYellow!70
      [
        {\citeleaf{%
          MPS-Prover~\citep{liang2025mps},
          3D-Prover~\citep{lamont2025d3prover},
          BFS-Prover~\citep{xin2025bfsprover},
          Bourbaki~\citep{zimmer2025bourbaki},
          HTPS~\citep{lample2022hypertree},
          AlphaProof~\citep{hubert_olympiad-level_2025}%
        }},
        fill=ntpYellow!15
      ]
    ]
    [
      {\textbf{Verifier-Guided Repair}},
      fill=ntpYellow!70
      [
        {\citeleaf{%
          Baldur~\citep{first2023baldur},
          HybridProver~\citep{hu2025hybridprover},
          Lyra~\citep{zheng2024lyra},
          ReForm~\citep{chen2025reform},
          ImProver~\citep{ahuja2025improver},
          ProofOptimizer~\citep{gu2026proofoptimizer}%
        }},
        fill=ntpYellow!15
      ]
    ]
  ]
  [
    {\textbf{Stateful / Agentic Workflows}},
    fill=ntpGreen
    [
      {\textbf{Persistent State and Libraries}},
      fill=ntpGreen!70
      [
        {\citeleaf{%
          FormalEvolve~\citep{lu2026formalevolve},
          LEAP~\citep{kung2026leap},
          Goedel-Architect~\citep{chung2026goedelarchitect},
          DreamProver~\citep{zhang2026dreamprover},
          ProofEvolve~\citep{ye2026proofevolve},
          LeanAgent~\citep{kumarappan2025leanagent},
          LEGO-Prover~\citep{wang2024lego}%
        }},
        fill=ntpGreen!15
      ]
    ]
    [
      {\textbf{Tool Use and Interfaces}},
      fill=ntpGreen!70
      [
        {\citeleaf{%
          LeanDojo~\citep{yang2023leandojo},
          CoqGym~\citep{yang2019coqgym},
          Seed-Prover~1.5~\citep{chen2025seedprover15masteringundergraduatelevel},
          OProver~\citep{ma2026oprover},
          Numina-Lean-Agent~\citep{liu2026numina},
          PALM~\citep{lu2024palm},
          ImProver~\citep{ahuja2025improver}%
        }},
        fill=ntpGreen!15
      ]
    ]
    [
      {\textbf{Multi-Agent Coordination}},
      fill=ntpGreen!70
      [
        {\citeleaf{%
          APOLLO~\citep{ospanov2025apollo},
          ProverAgent~\citep{baba2025prover},
          MASA~\citep{zhang2025masa},
          Aristotle~\citep{achim2025aristotle}%
        }},
        fill=ntpGreen!15
      ]
    ]
    [
      {\textbf{Knowledge Promotion \& Reuse}},
      fill=ntpGreen!70
      [
        {\citeleaf{%
          DreamProver~\citep{zhang2026dreamprover},
          ProofEvolve~\citep{ye2026proofevolve},
          LeanAgent~\citep{kumarappan2025leanagent},
          LEGO-Prover~\citep{wang2024lego}%
        }},
        fill=ntpGreen!15
      ]
    ]
  ]
]
\end{forest}%
}
\caption{\textbf{Three-axis methodology map for LLM-based formal theorem
proving.} Training strategies describe how model components are fitted;
inference-time mechanisms describe how proof candidates are generated,
retrieved, ranked, searched, and repaired; and stateful or agentic workflows
describe how state, tools, feedback, and delegation are coordinated across
calls. The axes are non-exclusive, and the cited systems are illustrative
rather than exhaustive. Retrieval includes formal premises and examples used
in statement translation or prompting.}
\label{fig:ntp-taxonomy-outline}
\end{figure*}

\Needspace{9\baselineskip}
\subsection{Training Strategies}
\label{sec:training-strategies}

\subsubsection{Standard Training Paradigms}
\label{sec:standard-paradigms}

\textbf{\revadd{Supervised fine-tuning.}}
\revadd{Many LLM-based provers begin with supervised fine-tuning (SFT) on
human-written or generated formal artifacts
~\citep{xin2024deepseekprover,lin2025goedel,wang2025kimina,
wang2024theoremllama}. DeepSeek-Prover creates data by autoformalizing
competition problems, filtering candidate statements, and generating checked
proofs, whereas Goedel-Prover combines translated statements with iterative
proof search~\citep{xin2024deepseekprover,lin2025goedel}. Alchemy instead
mutates existing library material symbolically~\citep{wu2025alchemy}. These
pipelines enlarge the available corpus, but kernel acceptance of a proof does
not by itself establish that an autoformalized statement preserves the source
problem; Section~\ref{subsec:data_scarcity} returns to fidelity and provenance.}

\begin{revisionblock}
\textbf{Reinforcement learning (RL).} In contrast to SFT on retained artifacts,
RL updates a policy from rewards assigned to sampled behavior. For formal
provers, the important distinction is the source and granularity of that
feedback. DeepSeek-Prover-V1.5 and
Leanabell-Prover use proof-assistant outcomes as rewards
~\citep{xin2024deepseekproverv15,zhang2025leanabell}, and
Leanabell-Prover-V2 integrates verifier feedback into reinforcement learning
~\citep{ji2025leanabellv2}. DeepSeek-Prover-V2 constructs
decomposition-structured cold-start trajectories before reinforcement learning
~\citep{ren2025deepseekproverv2}. Goedel-Prover-V2 combines expert iteration and
RL with a separate verifier-guided self-correction mode; self-correction is not
itself a reward definition~\citep{lin2025goedelv2}. Process-verified RL uses
Lean elaboration to label locally valid tactics and the earliest failure,
providing denser process rewards~\citep{kim2026processverified}.

FlowBalance supplies a boundary case: it calibrates on-policy self-guidance with
terminal verifier rewards for informal answers rather than proof-assistant
acceptance of formal artifacts
~\citep{HuangEtAl2026FlowBalanceArXiv260903241}. Reward definition, interaction
protocol, and compute budget must therefore be reported before gains can be
attributed to RL rather than to the surrounding pipeline.
\end{revisionblock}



\subsubsection{\revadd{Search-Coupled Training}}
\label{sec:task-specific-strategies}
\revadd{\textbf{Search-in-the-loop training.} In expert iteration, checked
search output becomes training data, so search quality and data selection are
coupled. A pre-LLM precedent trained neural guidance to rank unprocessed clauses
for processing
in the first-order prover \textsc{E}
~\citep{loos2017deep,schulz2002brainiac}. DeepSeek-Prover and HunyuanProver
combine verifier-filtered synthetic-data construction with proof search,
although the coupling between those components differs
~\citep{xin2024deepseekprover,li2024hunyuan}. HTPS updates its models from
earlier proof searches, while InternLM2.5-StepProver uses a critic both to
select training problems and to guide proof search
~\citep{lample2022hypertree,wu2024internlm}. In the LLM-based systems, checked
search outputs can become training objects; purely inference-time allocation
mechanisms are discussed below.}
\Needspace{2\baselineskip}

\subsection{\revadd{Inference-Time Search and Adaptation}}

\begin{revisionblock}
Search may construct training data, allocate inference-time compute, or update a
persistent library; these roles can occur in one system. At inference time, the
central question is what state is expanded and how candidates are ranked.
MPS-Prover combines a learned critic with hand-designed heuristic rules
~\citep{liang2025mps}; 3D-Prover filters tactics for predicted quality and
diversity~\citep{lamont2025d3prover}; and BFS-Prover length-normalizes
best-first search to favor deeper paths~\citep{xin2025bfsprover}. AlphaProof
couples policy and value learning to Monte Carlo tree search
~\citep{hubert_olympiad-level_2025}. At IMO~2024, the reported 28/42 total came
from the combined AlphaProof--AlphaGeometry~2 system, not AlphaProof alone.
Comparisons among these methods require matched accounting of sampled
candidates, node expansions, verifier calls, wall-clock time, and accelerator
use. 
\end{revisionblock}

\begin{revisionblock}
\textbf{Verifier-guided repair.} Proof-assistant feedback can drive proof
repair, statement repair, or optimization rather than only final acceptance.
Baldur repairs a generated whole proof from verifier diagnostics, whereas
HybridProver extracts a sketch from a whole-proof candidate and completes it
step by step~\citep{first2023baldur,hu2025hybridprover}. Lyra separates tool
correction from conjecture correction driven by prover errors
~\citep{zheng2024lyra}, while ReForm applies a revise-and-check loop to
autoformalized statements~\citep{chen2025reform}. ImProver and ProofOptimizer
instead simplify or refactor proof code in response to the formal environment
~\citep{ahuja2025improver,gu2026proofoptimizer}. These systems revise different
objects, and compilation of a repaired statement must still be distinguished
from semantic fidelity.
\end{revisionblock}

\begin{revisionblock}
Recent systems expose different temporal scopes of retained state. FormalEvolve
keeps a per-problem candidate archive~\citep{lu2026formalevolve}; LEAP and
Goedel-Architect maintain evolving proof or dependency blueprints within a
problem~\citep{kung2026leap,chung2026goedelarchitect}. DreamProver consolidates
lemmas across problems, while ProofEvolve promotes kernel-checked schemas into a
cross-problem library~\citep{zhang2026dreamprover,ye2026proofevolve}. Earlier
systems expose related boundaries: LeanAgent continually updates a reusable
store across tasks, while LEGO-Prover extracts reusable subproofs
~\citep{kumarappan2025leanagent,wang2024lego}. FormalEvolve's semantic gate is an
LLM-based judgment supplemented by reported manual faithfulness audits, not
kernel-certified equivalence. The operative
distinction is what persists after a call and what evidence licenses its reuse;
Sections~\ref{sec:sep2026-update} and~\ref{sec:literature-map-2026} compare these
boundaries in more detail.
\end{revisionblock}



\begin{revisionblock}
\textbf{Planning and theorem decomposition.} Decomposition is useful only when
the intermediate obligations are faithful and checkable. DSP maps an informal
draft to a formal sketch whose remaining subgoals are discharged by an
automated prover~\citep{jiang2023draftsketch}. \citet{cao2025reviving} revisit
this pattern with contemporary reasoning models.
DeepSeek-Prover-V2 instead uses recursive subgoal decomposition to construct
cold-start data and formal proofs~\citep{ren2025deepseekproverv2}.
DeepTheorem concerns informal natural-language proof generation and is
therefore a neighboring comparison, not evidence of machine-checked proof
production~\citep{zhang2025deeptheorem}. Across formal systems, a plan should be
evaluated by the checked obligations it induces, not by fluency alone.
\end{revisionblock}

\textbf{\revadd{Premise retrieval.}}
\revadd{Retrieval can supply formal declarations, natural-language exposition,
or previously solved proof fragments; these objects have different evidentiary
roles. ReProver selects accessible Lean premises, whereas COPRA retrieves
lemmas inside a stateful backtracking loop
~\citep{yang2023leandojo,thakur2024context}. HILBERT combines a semantic theorem
retriever with recursive decomposition~\citep{varambally2025hilbert}. The
LemmaHead preprint instead retrieves published-text context before generating
Lean proofs~\citep{yang2025lemmahead}. Retrieval results should report the
corpus snapshot, admissible-premise boundary, context budget, and overlap
controls; persistent stores introduce the additional versioning questions
discussed in Section~\ref{sec:literature-map-2026}.}

\begin{revisionblock}
\textbf{Domain-specific contrast.} Two inequality systems illustrate
different neuro-symbolic inference designs without general-purpose agentic
orchestration. The ICLR 2025 system assigns scaling to symbolic methods and
rewriting to a language model, then combines symbolic goal filtering with
language-model ranking
~\citep{li2025olympiadinequalities}. AIPS uses value-curriculum learning on
generated data to guide mixed-reasoning search
~\citep{wei2025olympiadalgebraic}. Their reported results remain specific to
those tasks and evaluation protocols.
\end{revisionblock}

\Needspace{8\baselineskip}
\subsection{\revadd{Agentic Proof Systems}}\label{subsec:agentic}
\begin{figure}[h]
    \centering
    \includegraphics[width=\linewidth]{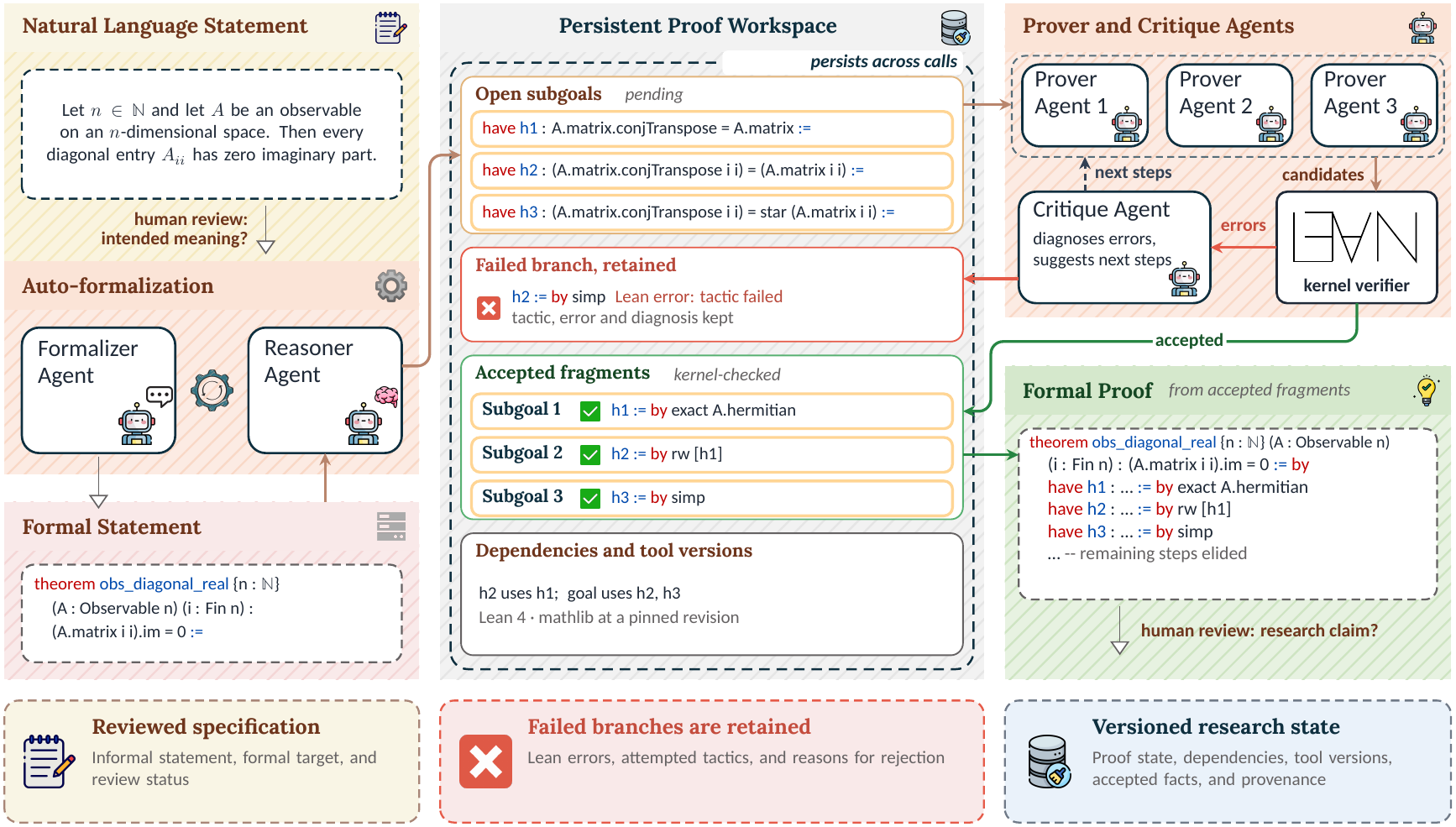}
    \caption{\textbf{Persistent agentic workflow for formal proof
    development.} A formalizer and a reasoning agent create and revise the
    formal statement. A versioned workspace stores subgoals, accepted proof
    fragments, failed branches, dependencies, and tool versions. Prover agents
    generate candidates, a critique agent diagnoses failures, and the kernel
    verifier determines formal acceptance. Human review remains necessary when
    the intended meaning or research claim is not captured by kernel checking.
    The displayed code is schematic and illustrates workspace organization
    rather than a complete kernel-checked Lean proof.}
    \label{fig:provisional-agentic-proving}
\end{figure}

\revadd{We use \emph{agentic} as a workflow label, not as a claim about a new
verifier. An agentic workflow preserves state across model and tool calls and
selects or schedules operations such as planning, retrieval, execution, repair,
and delegation; control may be learned or scaffold-defined. The checked
artifact and trusted base remain those of the underlying proof assistant.}
\revadd{Figure~\ref{fig:provisional-agentic-proving} makes the persistence
boundary explicit: accepted fragments, failed branches, dependencies, and tool
versions survive across calls.}

\revadd{Systems differ in the operations they orchestrate. ProverAgent combines
informal reasoning, formal proving, and lemma generation
~\citep{baba2025prover}; Aristotle adds a dedicated geometry solver to informal
lemma generation and Lean proof search~\citep{achim2025aristotle}. Two 2026
preprints broaden this design space: OProver couples retrieval, compiler
feedback, repair trajectories, and
iterative post-training~\citep{ma2026oprover}, whereas Numina-Lean-Agent places
a general-purpose coding agent over MCP-accessible Lean and retrieval tools
without task-specific prover training~\citep{liu2026numina}. These are distinct
pipelines, not instances of one controlled notion of agency.}

\revadd{LeanDojo and CoqGym expose proof states and actions on which such
controllers can operate~\citep{yang2023leandojo,yang2019coqgym}.
Seed-Prover~1.5 extends reinforcement learning across Lean and other tool
interactions~\citep{chen2025seedprover15masteringundergraduatelevel}, while the
Tencent system decouples informal reasoning from formal proving
~\citep{tencent_multiagent_IMO2025}. Because orchestration can increase
specialization, parallel sampling, and total inference cost simultaneously, its
contribution requires budget-matched ablations and explicit accounting of model
calls, tool calls, and verifier executions.}
\revadd{The reporting checklist in Section~\ref{sec:literature-map-2026}
provides a common disclosure frame for these otherwise heterogeneous systems.}

\Needspace{8\baselineskip}
\section{\revadd{Selected Benchmark Results and Research-Level Records}}
\label{sec:sota_benchmarks}

\revadd{Having described the main methodological components, we now turn to the
evidence used to assess them. This section first compares selected benchmark
records, then examines research-level evaluations and a dated observational
record of AI-related Erd\H{o}s-problem activity.}

\subsection{Formal Benchmark}

\revadd{miniF2F is a widely used benchmark for neural theorem
proving~\citep{zheng2022minif2f}. The Seed-Prover paper reports 243 of the 244
test problems solved (99.6\%) under its ``medium'' inference
setting~\citep{chen2025seed}. This near-ceiling, configuration-specific result
motivates harder evaluations; Table~\ref{tab:minif2f_sota} places it in
chronological context without treating heterogeneous budgets, checkpoints,
library versions, or formalizations as directly comparable.}

\begin{table}[h]
\centering
\begingroup
\ifshowrevisions\color{revisionblue}\fi
\fontsize{8}{8.7}\selectfont
\setlength{\tabcolsep}{2.2pt}
\renewcommand{\arraystretch}{0.98}
\begin{tabularx}{0.975\textwidth}{@{}>{\raggedright\arraybackslash}p{0.220\textwidth} >{\centering\arraybackslash}p{0.078\textwidth} >{\centering\arraybackslash}p{0.105\textwidth} >{\centering\arraybackslash}p{0.105\textwidth} X@{}}
\toprule
\rowcolor{gray!15} \textbf{System / configuration} & \shortstack[c]{\textbf{Backbone}\\\textbf{or size}} & \shortstack[c]{\textbf{Reported}\\\textbf{test result}} & \shortstack[c]{\textbf{First public}\\\textbf{date}} & \textbf{Technique / evaluation setting} \\
\midrule
Proof Artifact Co-training~\citep{han2021proof,zheng2022minif2f} & 837M & 29.2\% & Sep 2021 & \texttt{pass@8}; proof-artifact co-training and best-first search \\
HyperTree Proof Search~\citep{lample2022hypertree} & 600M & 41.0\% & May 2022 & \texttt{pass@64}; Monte Carlo tree search \\
ReProver~\citep{yang2023leandojo} & 299M & 26.5\% & Jun 2023 & \texttt{pass@1}; retrieval-augmented step-level proving \\
COPRA~\citep{thakur2024context} & --- & 30.7\% & Feb 2024 & GPT-4 Turbo, retrieval, and an LLM-generated informal proof; one attempt with up to 100 model queries and a 1,200-s timeout \\
InternLM2-StepProver~\citep{wu2024lean} & 7B & 54.5\% & Jul 2024 & Cumulative 64-pass result; step-level supervision \\
Lean-STaR~\citep{lin2024lean} & 7B & 46.3\% & Jul 2024 & \texttt{pass@64}; chain-of-thought SFT and expert iteration \\
TheoremLlama~\citep{wang2024theoremllama} & 8B & 33.6\% & Jul 2024 & Two-round union, 128 generations per round; curriculum training \\
DeepSeek-Prover-V1.5-RL~\citep{xin2024deepseekproverv15} & 7B & 63.5\% & Aug 2024 & 32 RMaxTS runs $\times$ 6,400 generations/run; 16 CoT + 16 non-CoT runs \\
Abel~\citep{gloeckle2024abel} & 8B & 41.3\% & Oct 2024 & Source notation $1\!\times\!128\!\times\!64$ (factor units unspecified); HyperTree search \\
Alchemy~\citep{wu2025alchemy} & 8B & 36.5\% & Oct 2024 & 1 attempt $\times$ 32 tactics/step $\times$ 100 steps; best-first search; 10-min limit \\
InternLM2.5-StepProver~\citep{wu2024internlm} & 7B & 65.9\% & Oct 2024 & 256 passes $\times$ 32 states $\times$ 600 expansions; best-first/critic-guided split \\
HunyuanProver-7B~\citep{li2024hunyuan} & 7B & 68.4\% & Dec 2024 & 600 passes $\times$ beam 8 $\times$ 400 iterations; best-first search + distance critic \\
BFS-Prover~\citep{xin2025bfsprover} & 7B & 72.95\% & Feb 2025 & Cumulative over $\alpha\!\in\!\{0,0.5,1\}$; one fixed configuration reports $70.83\!\pm\!0.89\%$ \\
Goedel-Prover-SFT~\citep{lin2025goedel} & 7B & 64.7\% & Feb 2025 & $4\!\times\!6{,}400=25{,}600$ whole-proof generations; autoformalization and expert iteration \\
STP~\citep{dong2025stp} & 7B & 67.6\% & Mar 2025 & \texttt{pass@25,600}; iterative self-play \\
DeepSeek-Prover-V2~\citep{ren2025deepseekproverv2} & 671B & 88.9\% & Apr 2025 & \texttt{pass@8,192}; recursive subgoal decomposition for cold-start data, followed by RL \\
Kimina-Prover-Preview~\citep{wang2025kimina} & 72B & 80.7\% & Apr 2025 & \texttt{pass@8,192}; formal-reasoning model with RL \\
Leanabell-Prover-GD-RL~\citep{zhang2025leanabell} & 7B & 61.1\% & Apr 2025 & \texttt{pass@128}; cognitive-behavior SFT and RL \\
DSP+~\citep{cao2025reviving} & --- & 83.6\% & Jun 2025 & Cumulative union across multiple draft--sketch--prove configurations \\
Kimina-Prover-72B + TTRL~\citep{KiminaTeam2025TTRL} & 72B & 92.2\% & 10 Jul 2025 & Test-Time Reinforcement Learning; adaptive lemma search; estimated pass upper bound $\sim$42,000 \\
Leanabell-Prover-V2-DS~\citep{ji2025leanabellv2} & 7B & 78.2\% & 11 Jul 2025 & \texttt{pass@128}; verifier-integrated SFT and RL \\
Delta-Prover~\citep{zhou2025decomposition} & \shortstack[c]{Gemini\\2.5 Pro\\05-06} & 95.9\% & 21 Jul 2025 & \texttt{pass@16,384}; reflective decomposition and repair \\
Seed-Prover~\citep{chen2025seed} & --- & 99.6\% & 31 Jul 2025 & ``Medium'': outer refinement + $8\!\times\!8$ light inner refinement; no fixed total disclosed \\
Goedel-Prover-V2-32B~\citep{lin2025goedelv2} & 32B & 92.6\% & 5 Aug 2025 & \texttt{pass@1,024}; synthesis, self-correction, and repair \\
HILBERT (Gemini 2.5 Pro + Goedel-Prover-V2-32B)~\citep{varambally2025hilbert} & Hybrid & 99.2\% & Sep 2025 & Single configuration; decomposition subset averages 548 reasoner + 391 prover calls; maxima 4.5K reasoner and 11.3K total/sample \\
ProverAgent ensemble~\citep{baba2025prover} & 8B + 7B & 88.1\% & 29 Sep 2025 & 260 prover calls; v5 (17 Feb 2026) additionally accounts for 84 auxiliary calls (344 total) \\
3D-Prover~\citep{lamont2025d3prover} & 7B & 53.1\% & Oct 2025 & Six searches; per node 128 proposals $\rightarrow\leq64$ unique tactics; DPP $K=8$ \\
\bottomrule
\end{tabularx}
\vspace{0.4em}
\caption{\revadd{Selected source-reported results on the 244-problem miniF2F
test split; percentages are transcribed rather than produced by a common rerun.
Rows are ordered by the first public appearance of the displayed result, at the
date precision recorded here; remaining ties are alphabetical. This is neither
the release chronology of the underlying systems nor a score ranking. Each row
reproduces one reported result, which may itself pool rounds or configurations
as stated. A dash means that the source reports no single applicable parameter
count.}}
\label{tab:minif2f_sota}
\endgroup
\end{table}

\Needspace{8\baselineskip}
\revadd{Each budget is reported in the source's native units. For whole-proof
sampling, \texttt{pass@N} typically denotes the fraction of benchmark problems
for which at least one of at most $N$ complete generated proofs is accepted.
Tree-search budgets may instead count runs, passes, states, candidates per
expansion, or expansion iterations and generally cannot be converted to a
common \texttt{pass@N}. RMaxTS means RMax applied to Tree Search, an
intrinsic-reward Monte Carlo tree-search method; Hunyuan's ``BFS'' means
best-first, not breadth-first, search; and DPP means determinantal point
process. A cumulative union pools configurations rather than estimating a
single fixed-budget pass rate. Abel does not define the units of all three
factors in its displayed budget. The displayed Kimina Test-Time Reinforcement
Learning (TTRL), Seed-Prover, and HILBERT configurations use branch-dependent
sequential calls and do not map cleanly to one fixed \texttt{pass@N}. The rows
are a chronological record, not a controlled ranking: checkpoints,
formalizations, Lean/mathlib versions, search budgets, and retry policies
differ. Fixed-budget reliability and transfer to other libraries or
research-level tasks remain separate questions.}

\begin{revisionblock}
The near-ceiling results in Table~\ref{tab:minif2f_sota} should therefore be
read as evidence that the standard miniF2F split has limited headroom under
high-budget evaluation, not as evidence that formal theorem proving is broadly
solved. FormalProofBench moves toward advanced undergraduate and graduate
mathematics, while MA-ProofBench isolates mathematical analysis at
undergraduate and PhD-qualifying levels
~\citep{ravi2026formalproofbench,pu2026maproofbench}. Their reported results are
not numerically comparable with miniF2F because the problem distributions,
access conditions, interaction protocols, and inference budgets differ. Taken
together, these benchmarks show why progress should be assessed across domain,
difficulty, library context, and resource budget rather than by a single pass
rate.
\end{revisionblock}

\Needspace{8\baselineskip}
\subsection{\revadd{Informal Benchmark}}
\label{subsec:first-proof}

\begin{revisionblock}
Unlike contest benchmarks, the second First Proof batch used ten research-level
problems for which the contributing mathematicians held complete solutions that
had not been made public. It evaluated four systems on all ten problems. An API
failure produced no output for one system--problem pair, so the released corpus
contains 39 submissions; each has at least two domain-referee reports
~\citep{abouzaid2026secondbatch,FirstProof2026Batch2Repository}. Across all four
systems, seven of the ten problems had at least one submission assigned one of
the two highest editorial categories, ``essentially flawless'' or ``requires
minor revisions''~\citep{abouzaid2026secondbatch}. Because this is a union over
systems, it is not a 70\% score for any one system; nor does the ``requires minor
revisions'' category by itself mean that a submission was publication-ready.
The released outputs,
configurations,
reported costs, and referee rationales support problem-level inspection. Nine
survey authors are listed as developers or principal investigators of
System~B; the post-conclusion \emph{Disclosure of Author Involvement, Funding,
and AI Assistance} section identifies this conflict of interest.
\end{revisionblock}

\begin{revisionblock}
First Proof's third batch, released on 14 October 2026, evaluated five systems
on ten research-level problems whose contributors held complete, unpublished
solutions of at most twelve pages. The organizers allowed 24 hours for each
system's ten-problem run, with no interactive human assistance. Thirty domain experts
reviewed anonymized submissions, now scoring mathematical correctness,
exposition, and attribution separately
~\citep{abouzaid2026thirdbatch,FirstProof2026Batch3Repository}. Incorrect
solutions could receive negative scores, while blank submissions and
acknowledged partial results received zero on correctness. This design makes
the burden of reading an incorrect proof part of the evaluation.

Using the report's threshold of a mean referee correctness score of at least
$0.5$, ProofCouncil, GPT-6 Astra, MerLean, Siege, and Humanize passed
$9$, $7$, $4$, $3$, and $0$ of the ten problems, respectively
~\citep{abouzaid2026thirdbatch}. That threshold includes minor, readily
correctable errors; it is not a count of flawless proofs. ProofCouncil received
full exposition marks on four submissions, illustrating why correctness and
readability remain separate outcomes. The reported totals of \$1,888 for
ProofCouncil and \$36 for the single-call-per-problem Astra configuration
describe those runs, with different models, orchestration, and tool use;
they do not isolate the causal effect of a harness. Released code, raw outputs,
cost records, and referee reports support inspection of these distinctions
~\citep{FirstProof2026Batch3Repository}. Because both the problems and scoring
rubric changed, the second and third batches should not be pooled into a
single success rate.
\end{revisionblock}

\begin{revisionblock}

\section{\revadd{From Theorem Proving to Research-Scale Mathematics: Recent Public Records}}
\label{sec:sep2026-update}

Recent progress in AI for mathematics is not adequately described by benchmark
accuracy alone. The unit of work has expanded from isolated theorem statements
to repositories, multi-file formalization projects, persistent agent workflows,
and open-ended research claims. These settings differ not only in difficulty
but also in what counts as evidence. For a benchmark item, an accepted proof
term may suffice as the primary reported outcome; a project-scale formalization
additionally requires
dependency control, integration, and semantic correspondence; an original
research claim further requires scrutiny of novelty, attribution, exposition,
and mathematical significance.

This section examines selected public records available through
\textbf{\literaturecutoff}. We treat announcements, manuscripts, formal
repositories, interaction traces, and external review as distinct components of
an evidence package. The purpose is not to rank cases by prestige or autonomy,
but to identify what each record establishes and which conclusions remain
outside its evidentiary scope. Figure~\ref{fig:ai4math-2026-timeline} provides
historical context by tracing selected theorem-proving and formalization
systems from 2022 through \literaturecutoff.

\begin{figure}[h]
    \centering
    \includegraphics[width=0.99\textwidth]{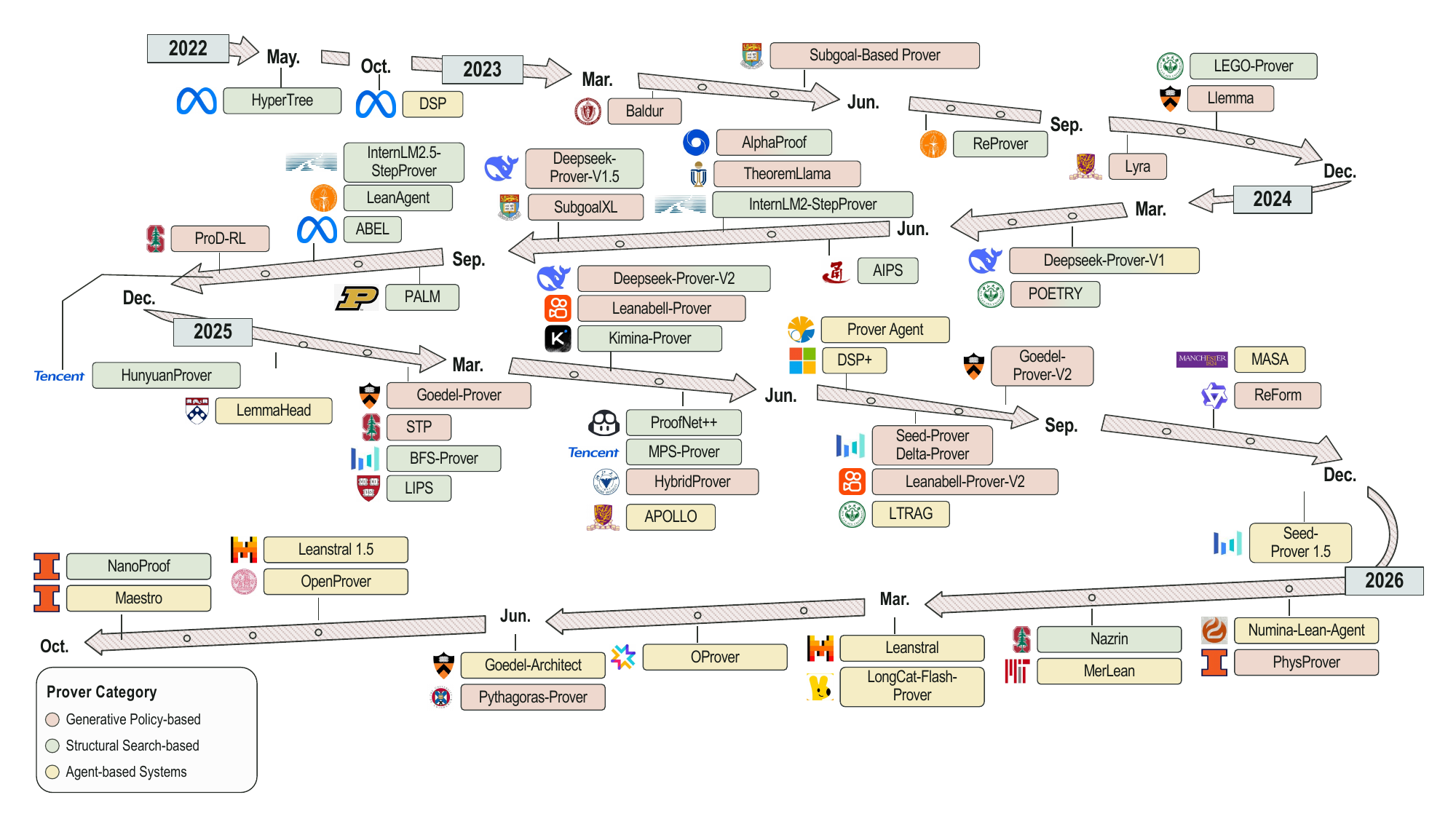}
    \caption{\revadd{\textbf{Selected AI-assisted theorem-proving and
    formalization systems, 2022--2026.} The source cutoff is
    \literaturecutoff. Positions indicate approximate public-release or
    first-preprint dates; spacing is not time-proportional. Colors indicate
    overlapping methodological emphases: generative policies, structural
    search, and agent-based workflows. The figure includes systems described
    in preprints; inclusion alone does not establish independent validation
    of reported results.}}
    \label{fig:ai4math-2026-timeline}
\end{figure}

The timeline complements the method taxonomy with dated system records.
Early examples include HyperTree and DSP~\citep{lample2022hypertree,jiang2023draftsketch},
Subgoal-Based Prover~\citep{zhao2023decomposing}, and
POETRY~\citep{wang2024proving}; the ProofNet++ node refers to its original
preprint~\citep{proofnetpp2025}.

The 2026 additions include Numina-Lean-Agent and PhysProver
~\citep{liu2026numina,zhang2026physprover}; Nazrin and MerLean
~\citep{aniva2026nazrin,ren2026merlean}; Leanstral and LongCat-Flash-Prover
~\citep{Mistral2026Leanstral,wang2026longcatflashprover};
OProver, Goedel-Architect, and Pythagoras-Prover
~\citep{ma2026oprover,chung2026goedelarchitect,leang2026pythagorasprover};
Leanstral~1.5 and OpenProver~\citep{Mistral2026Leanstral15,kripner2026openprover};
and NanoProof and Maestro~\citep{kripner2026nanoproof,li2026maestro}.
These systems provide technical context for the project-scale and open-ended
research records examined below.

\subsection{\revadd{A Shift in the Unit of Work}}

Recent systems increasingly preserve state across calls and across problems.
Process-verified reinforcement learning uses Lean-validated tactic steps and the
earliest failure location as reward signals~\citep{kim2026processverified};
DreamProver accumulates reusable lemmas, while ProofEvolve maintains proof DAGs
and promotes checked schemas into a cross-problem library
~\citep{zhang2026dreamprover,ye2026proofevolve}. LEAP coordinates an evolving
blueprint with compiler feedback, and QED and Prove2Me expose different planning
and shared-project layers~\citep{kung2026leap,an2026qed,chen2026prove2me}.
These designs cannot be compared solely by the number of final goals closed.
The retained object, the gate for reusing it, the reconstructibility of the
trace, and the location of human judgment all become part of the evaluation.

The evaluation unit has expanded in parallel. RLMEval samples tasks from active
formalization projects~\citep{poiroux2025rlmeval}; TCS-Bench derives contextual
tasks from theoretical-computer-science papers
~\citep{cohenaddad2026tcsbench}; TaoBench tests robustness under mathematically
equivalent but unfamiliar interfaces~\citep{taylor2026taobench}; and First
Proof links problems, submissions, and released referee reports
~\citep{abouzaid2026secondbatch}. These protocols measure different objects and
should not be collapsed into a single leaderboard.

\subsection{\revadd{Project-Scale Formalization}}

Autoformalization now ranges from dependency retrieval for statement
translation~\citep{wang2026ddr}, through proof-DAG construction
~\citep{cabral2026proofflow}, to multi-file developments
~\citep{wang2026m2f,zhang2026leanmarathon}. At project scale, success requires
shared definitions, dependency ordering, integration, and repair. A successful
build is necessary evidence for the formal artifact, but it does not by itself
establish correspondence with the source mathematics or long-term
maintainability.

Anthropic reported an eleven-day, multi-agent Claude formalization of Fermat's
Last Theorem using Prove2Me, with occasional high-level human direction, and
released the Lean repository~\citep{Anthropic2026FLT,Anthropic2026FLTReasoning,
Anthropic2026FLTArtifact}. This project formalizes an established theorem rather
than establishing a new mathematical result. Its pinned repository reports a
successful build and axiom audit, together with a secondary NanoDa check using
four disclosed patches that do not alter typing rules
~\citep{Anthropic2026FLTArtifact}. The record makes checker scope and the
artifact environment inspectable, while semantic audit, readability, and
maintenance remain separate questions.

Miller reports a different division of labor for the Vlasov mean-field
derivation: a mathematician scoped definitions, chose decompositions, and
triaged library gaps; an AI agent wrote the Lean development; and the paper
links a blueprint, logs, and a pinned repository
~\citep{Miller2026VlasovAIFormalization}. Together, these cases show why
project completion, semantic fidelity, engineering quality, and human effort
must be reported as distinct outcomes.

\subsection{\revadd{Open-Ended Mathematical Research}}

\subsubsection{\revadd{AI Activity on Erd\H{o}s Problems}}

\begin{revisionblock}
First Proof supplies a bounded problem set and released referee reports; the
community-curated Erd\H{o}s Problems AI wiki instead supplies an observational
register of heterogeneous activity~\citep{erdosAIWiki2026June}. The page
accessed on 4 September 2026 states that it is no longer updated and that its
latest data are as of 30 June 2026. Table~\ref{tab:erdos_contributions}
summarizes an author-prepared extract of those tables. Because the cited URL
does not pin an immutable revision, the counts are a dated record rather than a
permanently replayable snapshot or a cumulative growth series.

\begin{table}[h]
\ifshowrevisions\color{revisionblue}\fi
\centering
\small
\renewcommand{\arraystretch}{1.1}
\begin{tabularx}{\textwidth}{p{3.65cm} r r X}
\toprule
\rowcolor{gray!15} \textbf{Wiki category} & \textbf{Unique problems} & \textbf{Listed records} & \textbf{Operational meaning in the dated extract} \\
\midrule
AI standalone & 57 & 61 & Human involvement labeled ``non-significant''; comparable literature not identified. \\
AI alongside literature & 24 & 25 & Human involvement labeled ``non-significant''; comparable literature identified afterward. \\
AI building on literature & 37 & 38 & Human involvement labeled ``non-significant''; comparable literature known beforehand. \\
Human--AI collaboration & 116 & 132 & Human involvement labeled ``significant''; literature status varies. \\
Literature search & 77 & 84 & Recorded AI-assisted search for prior or related literature. \\
Formalization & 184 & 185 & Recorded formalization of a listed proof or result. \\
Rewriting & 9 & 10 & Recorded assistance with rewriting or exposition. \\
Computation & 3 & 3 & Recorded computational assistance. \\
\bottomrule
\end{tabularx}
\caption{\revadd{Author-computed category counts from the Erd\H{o}s Problems AI
wiki, accessed 4 September 2026; the page states that its latest data are as of
30 June 2026~\citep{erdosAIWiki2026June}. ``Unique problems'' deduplicates
problem identifiers within each category, whereas ``listed records'' counts
table rows. The operational-meaning column paraphrases the wiki;
``significant'' and ``non-significant'' are its labels rather than thresholds
defined by this survey.}}
\label{tab:erdos_contributions}
\end{table}

Both displayed counts are numerically larger for formalization, human--AI
collaboration, and literature search than for standalone AI activity. This
describes the composition of the curated page, not prevalence or success:
categories overlap, inclusion is selective, no denominator of attempted
problems is supplied, and the underlying register mixes full, partial,
incorrect, and unverified outcomes. Category assignments can also change when
prior work is found. Problem obscurity is not difficulty, and acceptance of a
proof for an encoded or weakened formalization does not by itself settle the
intended conjecture.

Beyond the wiki, research projects expose different units of evidence.
Feng et al.\ report AI-driven natural-language screening followed by human
expert evaluation of 700 conjectures labeled ``Open'' in Bloom's Erd\H{o}s
Problems database. They classify 13 as addressed: five as seemingly novel
autonomous solutions and eight as cases in which prior solutions were located
in the literature~\citep{feng2026semi}. The resulting $13/700$ ratio is not a
proof-success rate: its numerator combines solution and literature-retrieval
outcomes, and its denominator inherits database labels rather than representing
700 independently adjudicated open problems.
The LeanMarathon authors describe three runs as autonomous; across those runs,
the system formalized seven target theorems from two papers spanning four
Erd\H{o}s problems, with no \texttt{sorry} in the resulting Lean
artifacts~\citep{zhang2026leanmarathon}.

Separate mathematical papers disclose still different roles. Barreto et al.\
attribute the original Erd\H{o}s--Graham solution to Aletheia and describe the
remaining material as largely arising from human--AI interaction
~\citep{barreto2026irrationality}. Ma and Tang describe an AI assistant as an
early exploratory aid while stating that the authors independently verified
and wrote the final statements and proofs~\citep{ma2026erdos}. Lee and Seo
attribute key technical steps to a custom Gemini Deep Think research agent and
provide a later Lean formalization~\citep{lee2026independence}. Formal
Conjectures supplies versioned Lean~4 statements and frozen evaluation subsets
for research-level proof and disproof attempts
~\citep{firsching2026formalconjectures}. These records are not exchangeable
trials: natural-language screening, author disclosure, expert review, and
kernel checking support different claims. In particular, proof-assistant
acceptance checks an artifact against its encoded statement; it does not by
itself establish faithful informal interpretation or historical novelty. The
next section therefore treats target selection, artifacts, checking, and
provenance case by case.
\end{revisionblock}

\Needspace{12\baselineskip}

\subsubsection{\revadd{Navier--Stokes as a research-scale evidence package.}}
On 8 September 2026, OpenAI released a 166-page manuscript and a Lean
repository claiming a resolution of the Navier--Stokes existence and smoothness
problem~\citep{OpenAI2026NavierStokesAnnouncement,
OpenAI2026NavierStokesPaper,OpenAI2026NavierStokesLean}. For every positive
viscosity, the manuscript constructs a three-dimensional incompressible flow
that starts from rest, is driven by a smooth compactly supported force, and
develops unbounded velocity in finite time while retaining uniformly bounded
kinetic energy. The construction is presented as establishing alternatives C
and D in the official Clay problem formulation, respectively on
$\mathbb{R}^{3}$ and the periodic torus
~\citep{Fefferman2000NavierStokesProblem}.

OpenAI reports that the result emerged from a coordinated system involving on
the order of 10,000 concurrent agents. For the Navier--Stokes effort, the agents
exchanged approximately 2.7 million messages and generated about 130 billion
output tokens over 88 hours; Lean formalization and verification reportedly
required a further 17 hours using GPT-6 Astra
~\citep{OpenAI2026NavierStokesAnnouncement}. These figures document an
unusually large research workflow, but not a benchmark denominator or an
independently reproducible estimate of general mathematical capability. The
underlying research model was internal, so the public release supports direct
inspection of the manuscript and formal artifact more readily than replication
of the discovery process.

The case separates forms of validation that are often conflated. The Lean
repository supplies machine-checkable artifacts for encoded statements in a
specified software environment; the manuscript supplies the analytical
construction and its mathematical interpretation. Neither artifact alone
settles historical priority, correspondence between every formal statement and
its intended analytical meaning, or
the institutional status of the Millennium Prize. On 11 September 2026, the Clay
Mathematics Institute described the problem as ``apparently'' settled and
emphasized that evaluation and credit assignment would follow a deliberately
unhurried process~\citep{Clay2026NavierStokesAnnouncement}. This September
statement documents the institutional response at that date. We retain the
case as a formally accompanied proposed resolution and do not infer a later
review outcome from that announcement.

Figure~\ref{fig:navier-stokes-evidence-anatomy} maps the manuscript, formal
artifact, and selected public records as distinct parts of the evidence package.
The dated comparison includes Buckmaster's announcement of smooth-forced Euler
work with Alp\"oge alongside OpenAI's Navier--Stokes
release~\citep{buckmaster2026forcedblowupannouncement,OpenAI2026NavierStokesAnnouncement}.
These records concern different mathematical targets and do not, by their dates
alone, settle priority or credit.

\begin{figure}[h]
\centering
\includegraphics[width=\textwidth]{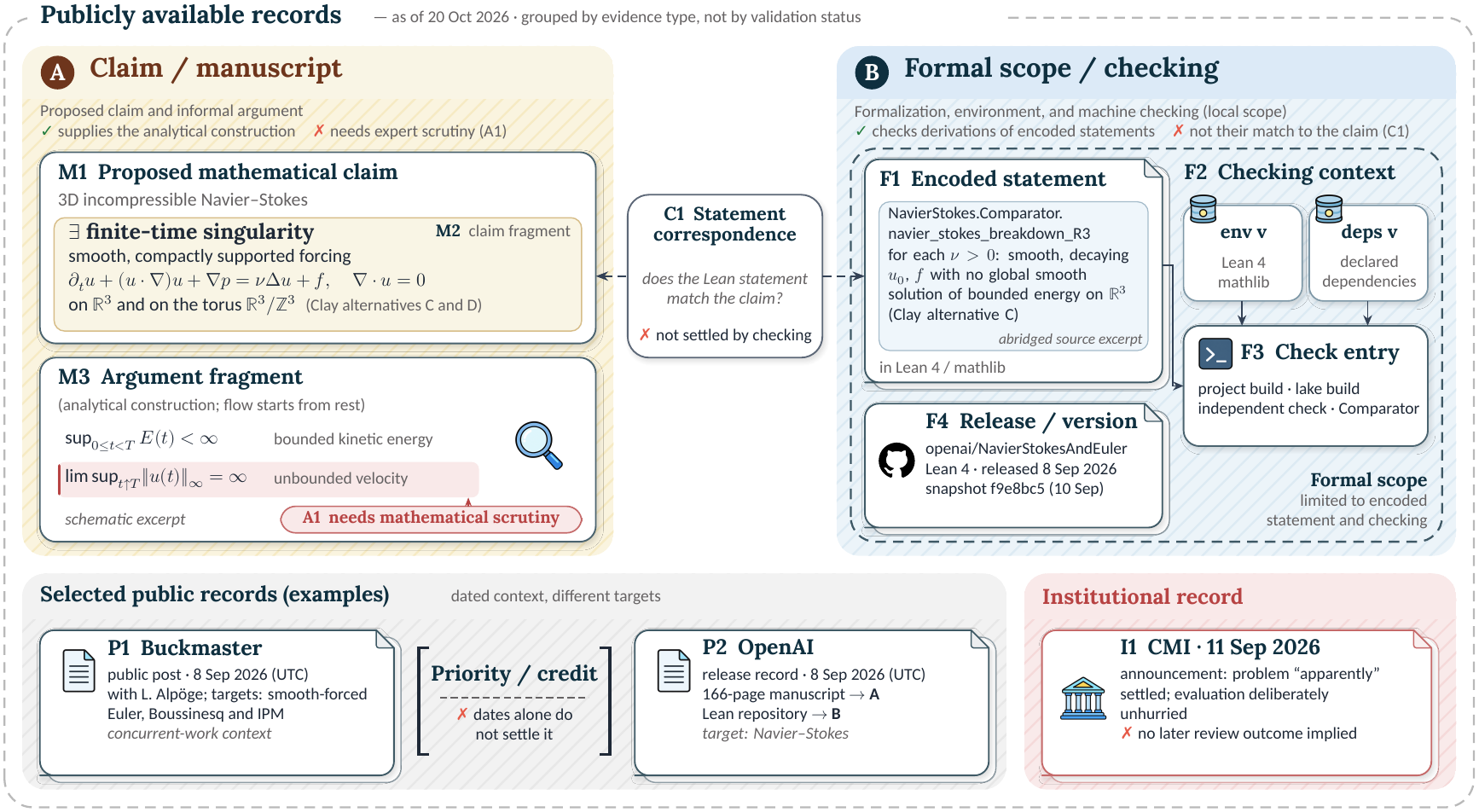}
\caption{\revadd{\textbf{Public evidence anatomy of the 2026 Navier--Stokes
proposed resolution.} The outer boundary groups selected public records
available by 20 October 2026, the survey's cutoff, and does not indicate a
common validation status. A, the
manuscript region presents the proposed three-dimensional incompressible
Navier--Stokes claim with smooth forcing, its whole-space and periodic settings,
and a schematic analytical
fragment~\citep{OpenAI2026NavierStokesPaper,Fefferman2000NavierStokesProblem}.
B, the formal region separates an abridged whole-space statement from its
environment, dependencies, checking entry points, and released
repository~\citep{OpenAI2026NavierStokesLean}. C1 marks the correspondence question
between manuscript and encoded statement; checking a formal derivation does not
itself resolve that question. Below, Buckmaster's announcement of related
smooth-forced Euler work with Alp\"oge is juxtaposed with OpenAI's release record
as concurrent-work context, not as an identical Navier--Stokes
claim~\citep{buckmaster2026forcedblowupannouncement,OpenAI2026NavierStokesAnnouncement}.
The CMI announcement represents a separate institutional
record~\citep{Clay2026NavierStokesAnnouncement}. Release dates do not adjudicate
priority, and public availability or checking instructions do not establish
independent mathematical acceptance or prize status. Mathematical fragments are
abbreviated and do not replace the full statements and hypotheses in the cited
sources.}}
\label{fig:navier-stokes-evidence-anatomy}
\end{figure}

Other 2026 cases expose complementary evidence profiles. The cycle-double-cover
record combines a disclosed prompt, proof manuscript, human exposition, and a
commit-pinned Lean project~\citep{OpenAI2026CDCPrompt,OpenAI2026CDCProof,
Oum2026CycleDoubleCover,OpenAI2026CDCLean}. The competing Crouzeix records pair
separately dated arguments with unusually explicit accounts of AI involvement,
but remained preprints at their stated release dates
~\citep{Jin2026CrouzeixPreprint,Jin2026CrouzeixArtifact,
LoristSchwenninger2026Crouzeix}. The zeta-zero result gives the unconditional
bound
\[
  \liminf_{T\to\infty}\frac{N_{0}^{s}(T)}{N(T)}
  \geq 0.6725007\ldots,
\]
where $N(T)$ counts nontrivial zeros with multiplicity and $N_{0}^{s}(T)$
counts simple zeros on the critical line. Its evidence package combines a
mathematical preprint, a reported discovery account, a Lean companion covering
a declared subset of statements, and a later independent preprint obtaining
the same constant~\citep{alpoge2026twothirds,Anthropic2026Zeta,
Anthropic2026ZetaLeanArtifact,Lamzouri2026Zeta}. Across these cases, formal
acceptance, independent mathematical argument, priority, and peer review are
different facts.

Provenance must also be read at the level of the full research chain. In the
Shannon-capacity sequence, an initial LLM-assisted construction was recast and
extended with exact-program checks and later strengthened by Lean-formalized
bounds~\citep{IttyEtAl2026ShannonOddCycles,Gao2026ShannonC7Recursive,
BuysEtAl2026ShannonLean}. In the 2026 probability sequence, author disclosures,
a Lean formalization of only the unit-slack case, and a later self-contained
proof of the stronger Samuels statement support different conclusions
~\citep{FuEtAl2026SmallDeviation,NieWei2026Feige,
WangZhang2026FeigeLeanArtifact,Ling2026Samuels}. Later extension or
formalization should therefore be presented as part of an evolving record, not
as an independent repetition of the original discovery.

\subsubsection{\revadd{Four Result Families in the OpenAI Mathematics Collection}}
\label{subsec:openai-october-families}

OpenAI's October mathematics release provides a collection-level example of
these distinctions. At the repository snapshot inspected for this survey,
the catalogue contains 719 manuscripts grouped into 372 result families.
Families can contain principal claims, supporting arguments, consequences,
and alternative proofs. OpenAI reports posing approximately 4,000 problems
and selecting and aggregating outputs for the catalogue
~\citep{OpenAI2026MathCollection}. Consequently, neither the manuscript count
nor the family count supplies a benchmark success rate. We examine four
families for their contrasting mathematical targets and public evidence;
Table~\ref{tab:openai-october-families} states the scope of each case.

\begin{table}[h]
\centering
\footnotesize
\renewcommand{\arraystretch}{1.18}
\ifshowrevisions
  \color{revisionblue}
  \arrayrulecolor{revisionblue}
  \captionsetup{labelfont={color=revisionblue},textfont={color=revisionblue}}
\fi
\begin{tabularx}{\textwidth}{>{\raggedright\arraybackslash}p{1.25cm}
                              >{\raggedright\arraybackslash}X
                              >{\raggedright\arraybackslash}X
                              >{\raggedright\arraybackslash}X}
\toprule
\rowcolor{gray!15}
\textbf{Family} & \textbf{Target claimed in the manuscripts} &
\textbf{Released evidence and stated scope} &
\textbf{Evaluation issue illustrated} \\
\midrule
102: Unique Games &
A Unique Games gap theorem and a direct proof of optimal Max-Cut
hardness~\citep{OpenAI2026UniqueGames,OpenAI2026MaxCutHardness}. &
Separate manuscripts and a Lean scope note identifying the encoded gap
reductions and approximation thresholds
~\citep{OpenAI2026UniqueGamesFormalization}. &
Related theorems and consequences require separate targets and dependency
records; counting papers does not count independent discoveries. \\
087: Mahler &
Symmetric and general convex-body volume-product bounds, including equality
cases~\citep{OpenAI2026MahlerSymmetric,OpenAI2026MahlerGeneral}. &
The Lean note includes selected geometric bounds and equality cases; the
family's functional inequalities lie outside those formalizations
~\citep{OpenAI2026MahlerFormalization}. &
Verification coverage must be recorded per statement, including equality
conditions and omitted extensions. \\
003: Quasi-Riemann &
A fixed zero-free half-plane $\operatorname{Re}(s)>7/8$ for the stated
$L$-functions~\citep{OpenAI2026QuasiRiemann}. &
The Lean note states the bound with pole exclusions and a uniform real-zero
gap; later applications in the paper are outside its scope
~\citep{OpenAI2026QuasiRiemannFormalization}. &
The exact region and hypotheses determine the claim; this is not the full
Riemann Hypothesis. \\
032: Hodge &
The rational Hodge conjecture for CM abelian varieties
~\citep{OpenAI2026CMHodge}. &
A revised manuscript and companion dependencies; the pinned catalogue has
no family-level Lean link~\citep{OpenAI2026MathCollection}. &
A restricted class of varieties must be distinguished from the general
Hodge conjecture, and manuscript dependencies remain part of review. \\
\bottomrule
\end{tabularx}
\caption{\revadd{\textbf{Four selected OpenAI result families and their
evidentiary roles.} The table describes claims and released documentation
at repository commit \texttt{1581d4cb} (12 October 2026), inspected by
\literaturecutoff. Formalization coverage is taken from the official scope
notes; this survey does not independently recheck their Lean developments
or certify the mathematical claims.}}
\label{tab:openai-october-families}
\end{table}

The collection also exposes limits on workflow comparison. OpenAI describes
an unreleased internal model and reports that the vast majority of results
used a common procedure, averaging three hours of ChatGPT Pro thinking
compute per result. The zero-free-region and CM Hodge work are stated
exceptions. Abridged reasoning summaries are available for the symmetric and
general Mahler results and for a basic-semidefinite-threshold hardness result
in Family~102~\citep{OpenAI2026MathCollection}. These summaries aid inspection
of selected arguments, while the reported average does not establish the
cost of each selected family or provide a replayable discovery workflow.

Version history adds evidence that a static list of claimed theorems would
miss. On 7 October, OpenAI withdrew three manuscripts concerning Weil
classes, Kuga--Satake correspondences, and products of K3 surfaces after a
shared sign error invalidated a supporting construction
~\citep{OpenAI2026MathHistory}. That notice does not withdraw the CM abelian
variety manuscript selected here; its 6 October edition separately records
updated companion references~\citep{OpenAI2026CMHodge}. The relevant unit of
assessment is therefore a dated claim together with its dependencies,
formalization coverage, corrections, and review record. These cases support
the evidence dimensions in Section~\ref{sec:literature-map-2026}, while the First
Proof third batch provides a complementary protocol with explicit tasks and
external scoring.

\subsection{\revadd{What the Recent Record Establishes}}

Taken together, the recent record shows a transition from local proof
production to longer-lived mathematical workflows. Systems now retrieve and
revise definitions, maintain state across many calls, construct multi-file
developments, and participate in open-ended research. The strongest public
records combine several kinds of evidence, but no common package has yet
emerged: some preserve formal artifacts without discovery traces; others
disclose model interactions without independent semantic audits; still others
provide expert review without a replayable workflow.

Accordingly, variation among public evidence packages is not incidental: it
determines what can be inferred from each record. A single label such as
``solved,'' ``verified,'' or ``autonomous'' suppresses distinctions among the
target statement, mathematical argument, formal encoding, computational
workflow, resource record, human intervention, and external review. The next
section examines platforms that support these workflows;
Section~\ref{sec:literature-map-2026} turns those distinctions into a comparative
framework.

\end{revisionblock}

\begin{revisionblock}

\section{\revadd{Platforms and Community Infrastructure for AI-Assisted Mathematics}}
\label{sec:community-platforms}

The project-scale workflows in Section~\ref{sec:sep2026-update} require ways to
retain mathematical context, coordinate contributions, and inspect intermediate
work. Research workspaces support these activities within a project, while
community platforms make problems, partial proofs, and completed formalizations
available to other contributors. We examine selected software and services
documented by \literaturecutoff, organized around interactive research, shared
formalization, and community repositories. Table~\ref{tab:community-platforms}
summarizes the work each platform retains and the access described in its source.

\begin{table}[t]
\centering
\small
\renewcommand{\arraystretch}{1.12}
\ifshowrevisions
  \color{revisionblue}
  \arrayrulecolor{revisionblue}
  \captionsetup{labelfont={color=revisionblue},textfont={color=revisionblue}}
\fi
\begin{tabularx}{\textwidth}{@{}>{\raggedright\arraybackslash}p{2.85cm}
                               >{\raggedright\arraybackslash}X
                               >{\raggedright\arraybackslash}X
                               >{\raggedright\arraybackslash}p{2.65cm}@{}}
\toprule
\rowcolor{gray!15}
\textbf{Platform} & \textbf{Main activity} &
\textbf{Retained or shared work} & \textbf{Documented access} \\
\midrule
Lea~\citep{LeaTeam2026Software} &
Interactive Lean formalization &
Proof versions, project memory, and blueprints &
Public source; local applications \\
Theo~\citep{moakhar2026theo} &
Research-paper autoformalization &
Definitions, statements, and Lean proofs &
Published workflow and formalization archive \\
AI co-mathematician~\citep{zheng2026aicomath} &
Exploratory mathematical research &
Hypotheses, failed approaches, and mathematical artifacts &
Limited initial release reported \\
Prove2Me~\citep{chen2026prove2me} &
Collaborative formalization missions &
Proof sketches, open subgoals, and reusable proofs &
Public platform \\
TheoremDB~\citep{Weiss2026TheoremDB} &
Shared proof-search memory &
Lean entries, proof states, and tactic outcomes &
Public read API; released client \\
Formal Conjectures~\citep{firsching2026formalconjectures} &
Research-problem formalization and curation &
Lean statements and community corrections &
Public repository \\
Palomar~\citep{Tao2026PalomarRegistry} &
Registration of formalized results &
Fixed repository snapshots and checked claims &
Public registry \\
\bottomrule
\end{tabularx}
\caption{\revadd{\textbf{Selected platforms and community infrastructure.}
The descriptions refer to the cited software snapshots and reports available
by \literaturecutoff. Access to a platform, its source, and its mathematical
outputs are reported separately where they differ.}}
\label{tab:community-platforms}
\end{table}

\subsection{\revadd{Interactive Research Workspaces}}

\href{https://vida-nyu.github.io/Lea/}{Lea} provides a Lean formalization agent
through a standalone web application and an Overleaf extension
~\citep{LeaTeam2026Software}. A mathematician can inspect successive proof
versions, edit the Lean code, and return it to the agent. Projects retain
instructions, accumulated memory, and a blueprint of dependencies among lemmas,
so a later session can continue an earlier argument. The Overleaf interface
connects labeled theorem blocks to formalization tasks. Its documentation also
distinguishes a successful Lean check from the additional kernel replay and
axiom checks performed by SafeVerify. These interfaces place proof development
and inspection alongside the mathematician's existing writing process.

\href{https://theo.quoin.workers.dev/}{Theo} organizes research-paper
autoformalization into stages for statements and proofs
~\citep{moakhar2026theo}. Its published workflow takes a paper in \LaTeX{} and
PDF form and produces Lean code, with an orchestrator that can revisit earlier
stages and accept human corrections during execution. When the paper needs
concepts absent from Mathlib, the system first constructs definitions and
uses auxiliary lemmas to test their intended properties. This exposes
definition choices before the main proof is attempted, giving researchers a
place to inspect how the source mathematics has been represented.

The AI co-mathematician extends the workspace approach to exploratory research
~\citep{zheng2026aicomath}. A coordinating agent delegates work on literature
search, computational exploration, and proof development while the researcher
can redirect ongoing tasks. The workspace retains hypotheses, unsuccessful
approaches, and mathematical artifacts across interactions. This supports
continuation of an evolving research project; the cited report describes a
limited initial release.

\subsection{\revadd{Shared Research and Formalization}}

Prove2Me makes a formalization project a shared undertaking
~\citep{chen2026prove2me}. Users launch ``missions'' with target statements, and
agents contribute proofs and intermediate results. Proof sketches expose
unresolved lemmas as separate tasks, allowing contributors to work on different
parts of an argument. The platform checks submitted proofs and provides a
mechanism for reusing established results in subsequent contributions. This
coordination supports the project-scale use discussed in
Section~\ref{sec:sep2026-update}. The mission design concentrates human auditing
on curated target statements, while agents develop supporting lemmas around
those targets.

TheoremDB provides a complementary form of shared memory for Lean proof search
~\citep{Weiss2026TheoremDB}. Its released client supports searching existing
entries, looking up proof states, and recording an attempted tactic together
with its outcome and execution cost. Public reads let a later agent consult
earlier work before repeating a search. The reusable record can therefore
include an individual proof attempt as well as a completed theorem.

\begin{figure}[t]
\centering
\includegraphics[width=\textwidth]{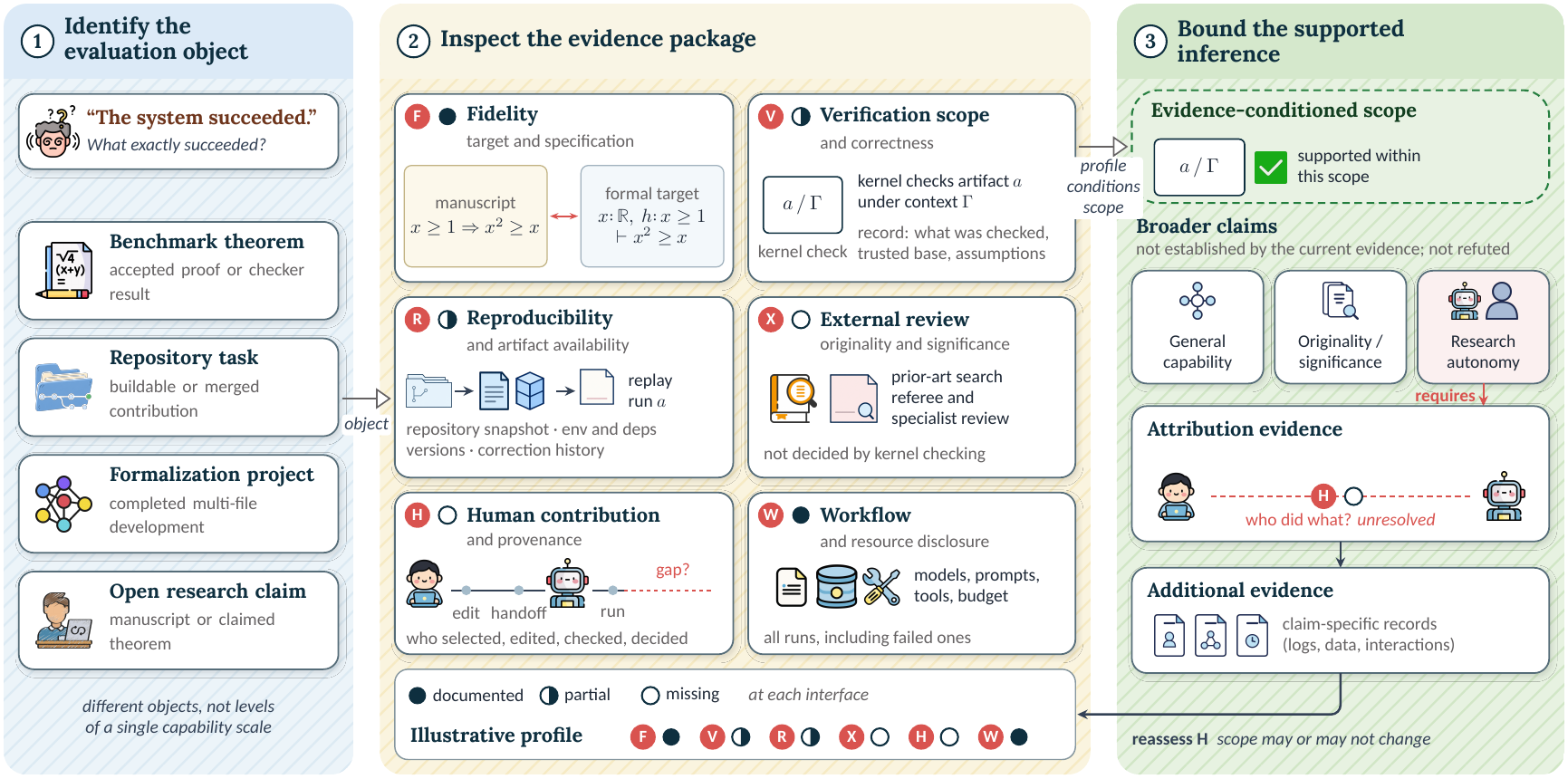}
\caption{\revadd{\textbf{Evidence-conditioned evaluation of AI-assisted
mathematical research.} Left, the evaluation object is identified as a benchmark
theorem, repository task, formalization project, or open research claim; these
are not levels of a single capability scale. Centre, shared objects expose six
evidence interfaces: target and specification fidelity (F), correctness and
verification scope (V), workflow and resource disclosure (W), artifact
availability and reproducibility (R), human contribution and provenance (H), and
originality, significance, and external review (X). The token $a$ identifies the
artifact, while $\Gamma$ denotes the context and assumptions relevant to checking.
Filled, half-filled, and open markers denote documented, partial, and missing
evidence at an interface. The mixed profile is illustrative, not an evaluation
of a named system or the Navier--Stokes case. Right, the profile conditions the
supported inferential scope. Broader candidate claims outside the boundary are
not established by the current evidence, rather than false or rejected. The
research-autonomy branch links unresolved attribution to additional
claim-specific evidence and reassessment of H. Reassessment may or may not
change the supported scope; additional evidence does not automatically establish
the broader claim.}}
\label{fig:evidence-conditioned-evaluation}
\end{figure}

\subsection{\revadd{Community Repositories and Registries}}

Formal Conjectures connects problem curation with proof attempts
~\citep{firsching2026formalconjectures}. Mathematicians contribute and clarify
research statements in Lean, while humans and AI systems attempt proofs or
disproofs. These attempts can also reveal errors in the formal statement and
prompt corrections to the shared collection. Beyond its benchmark role in
Section~\ref{sec:datasets_benchmarks}, the repository provides a common set of
precise targets around which contributors can organize mathematical work.

Palomar supports the dissemination and inspection of completed formalizations
~\citep{Tao2026PalomarRegistry}. It registers fixed snapshots of external Lean
repositories, pairing a readable formal statement with its proof and an
informal description. A mechanical check using Comparator establishes whether
the submitted proof proves the specified formal claim; a separate
model-assisted check compares that claim with its informal description.
Registration makes the checked version and the scope of these checks
identifiable. Expert assessment of novelty, significance, and the mathematical
argument remains a separate activity.

Project histories, shared proof sketches, curated statements, and registered
formalizations provide concrete material for examining how a result was
obtained and what can be reused. Section~\ref{sec:literature-map-2026} develops
the evidence dimensions needed to assess such records.

\end{revisionblock}


\begin{revisionblock}
\section{\revadd{Evaluating Research-Scale AI Mathematics}}
\label{sec:literature-map-2026}

AI-for-mathematics systems are often compared through a single headline
outcome: pass rate, accepted proof, completed project, or resolved problem.
These outcomes are not directly commensurate. They refer to different targets,
expose different failure modes, and rely on different combinations of machine
checking and human judgment. We therefore organize evaluation along two
complementary
structures: a spectrum describing the unit of mathematical work, and a set of
evidence dimensions describing the support available for a particular claim.
The framework synthesizes the literature available through
\textbf{\literaturecutoff}; it is not a ranking of systems or mathematical
results.

Figure~\ref{fig:evidence-conditioned-evaluation} summarizes the evaluation
logic: identify the object being assessed, inspect the relevant evidence
interfaces, and bound the inference supported by the available record.
Table~\ref{tab:ai4math-evaluation-spectrum} details the evaluation objects, while
Section~\ref{subsec:six-evidence-dimensions} defines the six evidence dimensions.


\subsection{\revadd{An Evaluation Spectrum}}

The evaluation spectrum runs from isolated benchmark problems through
repository-level proof engineering and project-scale formalization to
open-ended mathematical research. It is not a maturity ladder. A formally
checked benchmark submission may support a stronger local correctness claim
than an open-ended research manuscript, while the latter may demonstrate a
broader form of problem selection, conceptual discovery, or sustained project
work. Moving along the spectrum changes the evaluation object and increases the
number of interfaces that must be examined.

\begin{table}[h]
\centering
\small
\renewcommand{\arraystretch}{1.16}
\ifshowrevisions
  \color{revisionblue}
  \arrayrulecolor{revisionblue}
  \captionsetup{labelfont={color=revisionblue},textfont={color=revisionblue}}
\fi
\begin{tabularx}{\textwidth}{>{\raggedright\arraybackslash}p{2.45cm}
                              >{\raggedright\arraybackslash}p{2.45cm}
                              >{\raggedright\arraybackslash}X
                              >{\raggedright\arraybackslash}X}
\toprule
\rowcolor{gray!15}
\textbf{Evaluation object} & \textbf{Primary outcome} &
\textbf{Additional evidence needed} & \textbf{Inference not supported by the outcome alone} \\
\midrule
Benchmark theorem & Accepted proof or checker result &
Statement audit, benchmark version, denominator, search budget, and verifier configuration &
General research ability or robustness outside the benchmark \\
Repository task & Buildable or merged contribution &
Dependency state, tests, project conventions, review history, and maintainability assessment &
Semantic adequacy or architectural quality \\
Formalization project & Completed multi-file development &
Source--formal correspondence, integration history, axiom audit, and division of labor &
Autonomous mathematical understanding or general productivity \\
Open research claim & Manuscript, construction, or claimed theorem &
Prior-art search, expert scrutiny, provenance, formalization scope, resources, and external review &
Settled priority, significance, or field-wide capability \\
\bottomrule
\end{tabularx}
\caption{\revadd{\textbf{Evaluation spectrum for AI-assisted mathematics.}
The rows describe different evaluation objects rather than levels of a single
capability scale. Each primary outcome supports a local claim and requires
additional evidence before broader conclusions are drawn.}}
\label{tab:ai4math-evaluation-spectrum}
\end{table}

At the benchmark end, suites such as miniF2F and PutnamBench isolate theorem
statements and make proof acceptance comparatively direct
~\citep{zheng2022minif2f,tsoukalas2024putnam}. Repository-derived suites instead
inherit dependencies, style constraints, and incomplete project context.
SorryDB samples open theorem holes from working Lean repositories,
TheoremBench includes supporting subtheorems, and MathlibPR evaluates merge
readiness rather than goal closure alone
~\citep{letson2026sorrydb,pham2026theorembench,xie2026mathlibpr}. Project-scale
formalization adds definition design, cross-file integration, and source--formal
alignment; open research adds novelty, attribution, and significance. Results
can therefore be compared only after the evaluation object has been made
explicit.

\subsection{\revadd{Six Evidence Dimensions}}
\label{subsec:six-evidence-dimensions}

We distinguish six dimensions that recur across the records in
Section~\ref{sec:sep2026-update}. They are related but non-substitutable:
strength on one does not automatically compensate for missing evidence on
another.

\paragraph{\revadd{Target and specification fidelity.}}
This dimension asks whether the formal or operational task preserves the
intended mathematical problem. Autoformalization includes statement
translation, proof translation, informalization, and end-to-end formalization;
success on one task does not imply success on another
~\citep{mensfelt2025common}. Compilation establishes that a candidate
elaborates in a specified environment, not that it faithfully represents the
source. Existing studies use different proxies: semantic-equivalence judgments
~\citep{dai2026signalcoverage}, bidirectional provability
~\citep{mohammad2026faithfulnessgap}, human-calibrated model judges
~\citep{zhang2026beyondcompilation}, and forward--backward implication against
auxiliary ``shadows''~\citep{han2026shadowbench}. FaithformBench further shows
that an invalid source can be transformed into a provable but meaning-altered
target~\citep{CornishEtAl2026FaithformBenchArXiv260810916}. Reports should name
the alignment criterion, implication direction, abstention rule, and human
calibration rather than treating type-checking as semantic equivalence.
The October cases make the target boundary concrete: a fixed
$\operatorname{Re}(s)>7/8$ zero-free region and the rational Hodge conjecture
for CM abelian varieties each specify a narrower mathematical target than
the corresponding general conjecture
~\citep{OpenAI2026QuasiRiemann,OpenAI2026CMHodge}.

\paragraph{\revadd{Correctness and verification scope.}}
A correctness claim must identify the object checked, the trusted base, and the
assumptions that remain. A proof assistant's kernel checks an exact formal
object relative to its logic, axioms, implementation, and pinned dependencies.
External solvers, executable evaluators, model critics, and expert reviewers
support different claims. Reward-oracle audits show that compilation combined
with a source scan can admit hidden axioms such as \texttt{sorryAx}, while
kernel-level re-audits reject some outputs accepted by weaker filters
~\citep{vamshi2026rewardoracle}. Likewise, a Lean companion covering selected
lemmas does not certify every step of an accompanying prose proof. Verification
scope should be described positively---what was checked and under which
assumptions---rather than by the label ``formally verified'' alone.
For example, the Mahler scope note distinguishes geometric volume-product
bounds and equality cases from functional inequalities elsewhere in the same
family~\citep{OpenAI2026MahlerFormalization}. A family-level label would hide
that difference.

\paragraph{\revadd{Workflow and resource disclosure.}}
Agentic results depend on model versions, prompts, tools, search policies,
compute budgets, stopping rules, and treatment of failed branches. The
Atomic-to-Agentic benchmark demonstrates that systems with similar end-to-end
accuracy can have different profiles across planning, action, and feedback
~\citep{KuangEtAl2026AtomicAgenticMathArXiv260826950}. Persistent systems also
retain different objects---lemmas, proof graphs, blueprints, retrieval stores,
or shared workspaces~\citep{zhang2026dreamprover,ye2026proofevolve,
kung2026leap,chen2026prove2me}. A label such as ``multi-agent'' therefore does
not specify either the mechanism responsible for a result or the cost of
obtaining it. At minimum, reports should disclose the operational configuration
needed to interpret the claimed outcome.

\paragraph{\revadd{Artifact availability and reproducibility.}}
A paper, repository, formal environment, and interaction trace enable different
forms of inspection. Repositories support replay only to the extent that they
pin dependencies and preserve build instructions; reconstructing a discovery
workflow additionally requires prompts, trajectories, tool outputs, human
edits, and stopping conditions. Rollout-card audits found that many agent
evaluations omit failed, errored, or skipped runs and that alternative reporting
rules can materially change scores~\citep{masters2026rolloutcards}. Versioned
benchmarks matter for the same reason: Epoch's FrontierMath changelog records
substantial corrections and removals between releases
~\citep{EpochAI2026FrontierMathV2}. Reproducibility claims should therefore
identify both the artifact snapshot and the denominator from which reported
successes were selected.
The OpenAI collection's linked withdrawals further show why dependency
versions and correction histories belong in this record
~\citep{OpenAI2026MathHistory}.

\paragraph{\revadd{Human contribution and provenance.}}
The relevant record is not a binary choice between ``human'' and ``AI.'' It
includes who selected the problem, supplied definitions or references, chose
among branches, corrected semantics, checked prior art, wrote the exposition,
and accepted responsibility for the final claim. Aletheia's eigenweights case,
for example, separates reported machine generation of mathematical content
from human rewriting and introduction writing
~\citep{feng2026towards,feng2026eigenweights,
DeepMind2026AletheiaArtifacts}. Corpus-scale analyses of author disclosures
measure reporting under a stated protocol, not correctness or autonomy
~\citep{JinEtAl2026GoldRush}. Provenance should therefore be recorded as an
intervention history, with source-attributed autonomy labels retained only as
descriptions of what the cited authors claim.

\paragraph{\revadd{Originality, significance, and external review.}}
Kernel checking can validate a formal derivation but cannot by itself determine
whether the intended theorem is new, important, or correctly situated in the
literature. PriorProof tests whether proof techniques occur in a dated formal
corpus, but absence from that corpus is not a general novelty certificate
~\citep{somani2026priorproof}. Bibliographic resources such as Matlas and
TheoremSearch can make possible prior art searchable, while coverage,
extraction error, and semantic equivalence remain open issues
~\citep{ju2026matlas,alexander2026theoremsearch}. Referee reports, independent
proofs, specialist commentary, and institutional adjudication add evidence that
is not interchangeable with either retrieval or formal verification.
First Proof's third batch operationalizes part of this distinction by
scoring correctness, exposition, and attribution separately; the leading
system's nine correctness successes and four full exposition scores are
different measured outcomes~\citep{abouzaid2026thirdbatch}.

The profile in Figure~\ref{fig:evidence-conditioned-evaluation} is illustrative
and does not assess a named system or the Navier--Stokes case. Its states describe
evidence at audited interfaces, not capability scores or the truth of claims.
Broader claims outside the supported boundary remain unestablished by the current
record. The research-autonomy example shows how additional contribution and
provenance evidence can trigger reassessment without guaranteeing a change in
supported scope.

\subsection{\revadd{Comparing Heterogeneous Results}}

The six dimensions explain why superficially similar success claims can have
different meanings. A benchmark score requires a stable statement set, full
outcome accounting, and a declared budget. A repository contribution adds
integration and maintainability review. A formalization project adds an audit
of the relation between source mathematics and formal declarations. An
open-ended research result adds prior-art analysis, attribution, and external
mathematical scrutiny. These requirements accumulate because the number of
interfaces grows, not because later rows in
Table~\ref{tab:ai4math-evaluation-spectrum} are automatically more capable.

Benchmark auditing illustrates the interaction among dimensions. A corpus-scale
audit of five Lean benchmarks and their forks produced 4,833 recall-oriented
flags and mechanically certified 398 issues, including counterexamples,
vacuous theorems, and unsound axioms
~\citep{ammanamanchi2026faults,AmmanamanchiEtAl2026ATPCheckersArtifact}.
The larger number is a flag count, not a count of independently certified
mathematical errors. ProofJudge instead evaluates agreement with repository
revision history, while MathlibPR and expert reconstruction assess merge
readiness and project quality
~\citep{caldwell2026proofjudge,xie2026mathlibpr,ilin2026sorries}. These are
different evaluation targets rather than competing estimates of one score.

Research-level benchmarks broaden the source material without making their
outcomes exchangeable. RLMEval draws from active Lean Blueprint projects;
FormalTCS follows generated research claims through expert and formal checks;
TCS-Bench derives contextual tasks from published papers; and
AdvancedMathBench separates proof generation from expert-labeled error
detection~\citep{poiroux2025rlmeval,wang2026formaltcs,
cohenaddad2026tcsbench,kong2026advancedmathbench}. OEIS Open additionally
reports denominators and per-target budgets, ITPEval evaluates translation
among proof assistants, and MathAtlas emphasizes textbook entities and
dependency relations~\citep{Adamczewski2026OEISOpenArXiv260811941,
wu2026itpeval,patel2026mathatlas}. Each resource should be interpreted using the
claim supported by its own protocol.

Domain transfer and project scale introduce further specification boundaries.
MechGeo distinguishes direct proofs from cases where the system first produced
counterexamples to faulty formulations and experts repaired the targets
~\citep{shen2026mechgeo}. FormalScience and Lean4Physics examine scientific and
physics formalization, while FLARE, SOVER, and Agentic Proving study
optimization reformulations or program verification
~\citep{meadows2026formalscience,li2026lean4physics,robbins2026flare,
bhattacharyya2026sover,sosso2026agenticverification}. These systems differ in
domain assumptions, external tools, and the meaning of a verified outcome; a
single proof-completion metric would conceal those differences.

Large formalization projects should likewise be read as versioned engineering
records. Work on topology, De Giorgi--Nash--Moser theory, tensor networks,
economics and computation, and large Lean theories toward finite-group
classification provides different combinations of definitions, dependencies,
proof terms, and review metadata
~\citep{bryant2026munkres,urban2026130k,armstrong2026formalization,
lu2026tensor,garg2026econcslib,nie2026formatheoria}. Quantitative scale and
completion claims should name a dated release or immutable commit rather than
silently substituting a mutable repository for the reported snapshot.

\subsection{\revadd{A Reporting Checklist}}

For project-scale and open-ended AI4Math results, the following fields should be
disclosed or explicitly marked not applicable. The checklist operationalizes
the six dimensions above; outcome accounting is listed separately because the
denominator and excluded runs cut across evaluation, reproducibility, and
resource disclosure.
\begin{enumerate}[label=(\roman*), leftmargin=2.2em, itemsep=1.5pt, topsep=3pt,
                  parsep=0pt, partopsep=0pt]
    \item \textbf{Target and fidelity:} the exact informal target, any formal
    encoding, known differences between them, and the procedure used to assess
    correspondence.
    \item \textbf{Verification scope:} the prover or evaluator, dependency and
    library commits, axioms, trusted components, and the precise claims checked.
    \item \textbf{Workflow and resources:} available identifiers and settings
    for models, prompts, tools, compute, communication structure, stopping rules,
    and search budgets.
    \item \textbf{Outcome accounting:} attempted targets, failed or excluded
    branches, checker rejections, and the denominator used for aggregates.
    \item \textbf{Artifacts and reproducibility:} locations and versions of
    code, logs, proof objects, data, and build instructions, including known
    omissions.
    \item \textbf{Human contribution and responsibility:} material
    interventions, selection and correction decisions, exposition, and the
    people responsible for the final claim.
    \item \textbf{Originality and review:} prior-art procedures, independent
    arguments, expert- or peer-review status, priority claims, and unresolved
    objections.
\end{enumerate}

When lawful and practical, a replay bundle should additionally freeze exact
prompts, raw trajectories, tool outputs, and unsuccessful branches. Partial
disclosure must remain distinguishable from a reproducible run. The purpose of
the checklist is not to force every project into one benchmark, but to prevent a
single outcome from standing in for formal validity, semantic fidelity,
originality, exposition quality, cost, and human contribution at once.

\end{revisionblock}

\Needspace{11\baselineskip}
\section{Open Challenges and Future Directions}
\label{sec:challenges}


\revadd{The preceding evidence suggests that the central bottleneck is no
longer proof generation alone, but the conversion of local solver success into
durable mathematical knowledge. The five linked challenges below examine that
transition across data, representation, discovery, tool integration, and
human--AI collaboration.}

\subsection{\revadd{Formal-Mathematics Data and Evaluation}}
\label{subsec:data_scarcity}

\begin{revisionblock}
Scaling learning-based formal reasoning is constrained partly by the limited
supply of curated formal developments~\citep{weng2025autoformalization}.
Because formal and natural-language corpora use different units, tokenizers,
filtering rules, and dependency structures, simple order-of-magnitude
comparisons can mislead. Autoformalization offers one bridge, but exposes a
granularity mismatch: human exposition relies on implicit context and shared
conventions, whereas proof assistants require explicit declarations and side
conditions.
\end{revisionblock}

\begin{revisionblock}
A central challenge is to turn the semantic and artifact audits reviewed in
Sections~\ref{sec:sep2026-update} and~\ref{sec:literature-map-2026} into
versioned data pipelines. Generated and curated items need provenance,
assumptions, a pinned formal target, and a correction history. Benchmark audits
can expose meaning-altering discrepancies
~\citep{ospanov2025minif2fleanrevisitedreviewinglimitations}, while negative
controls test whether such alterations remain invalid
~\citep{CornishEtAl2026FaithformBenchArXiv260810916}. Project-scale case studies
further show that closing admitted goals is not the whole task
~\citep{ilin2026sorries}: proof simplification
~\citep{gu2026proofoptimizer} and merge readiness
~\citep{xie2026mathlibpr} are distinct evaluation targets.
\end{revisionblock}

\begin{figure}[h]
    \centering
    \includegraphics[width=0.98\linewidth]{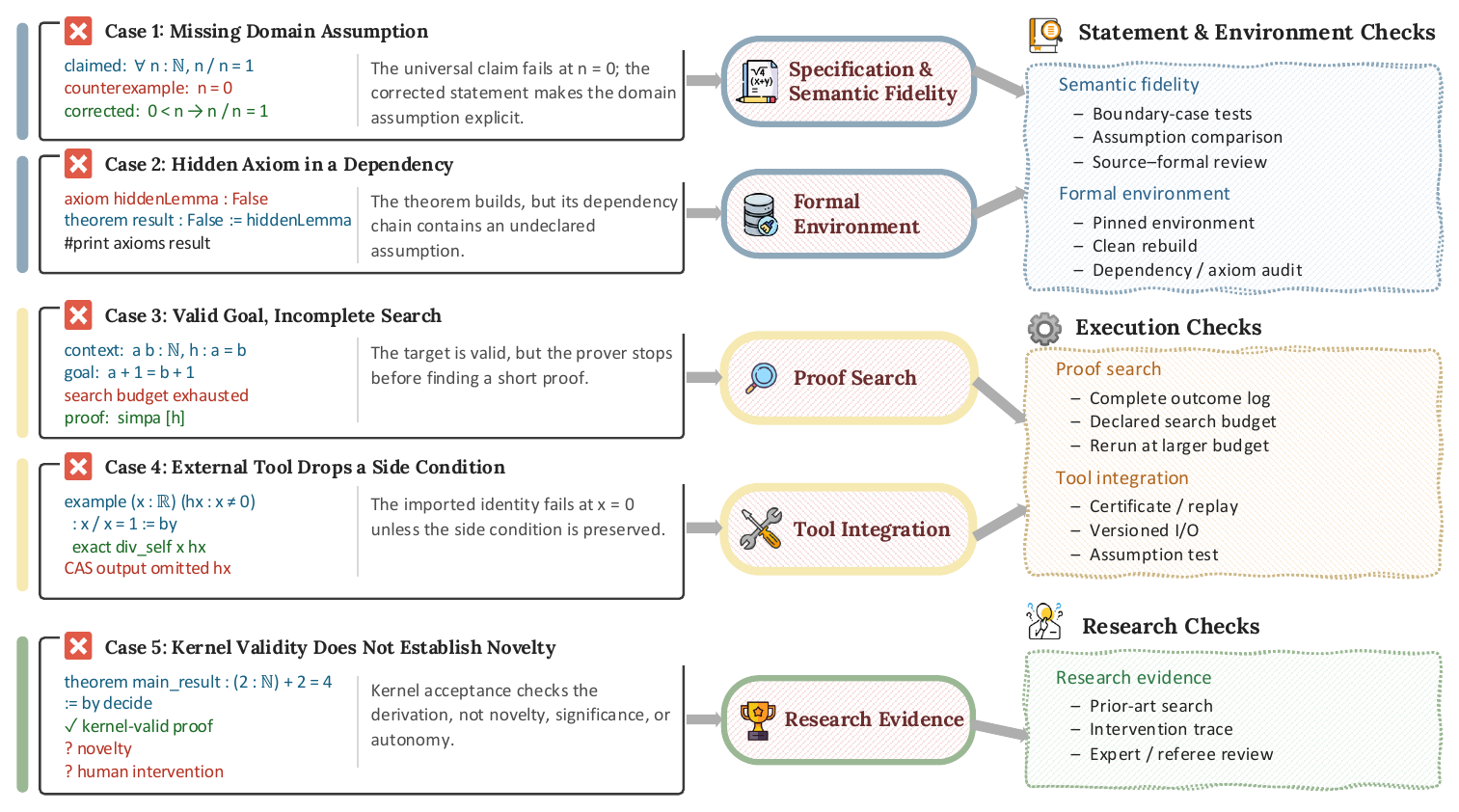}
    \caption{\textbf{Failure modes and corresponding checks in LLM-driven
    formal mathematics.} The examples distinguish failures in specification
    and semantic fidelity, the formal environment, proof search, tool
    integration, and research evidence. Each class requires a different form
    of diagnosis. Kernel acceptance establishes formal validity in the stated
    environment, but does not establish semantic fidelity, novelty,
    significance, or autonomous contribution.}
    \label{fig:provisional-failure-cases}
\end{figure}

\begin{revisionblock}
Recent systems expand formal-training corpora through lifted or augmented
natural-language--formal pairs, quality-filtered theorem synthesis, and
scaffolded proof-data synthesis
~\citep{liu2025atlas,wang2025qdtsynth,lin2025goedelv2}. The decisive experiment
is not whether a generator produces more compiling examples, but whether those
examples improve semantic fidelity, transfer, and maintainability on held-out
tasks under a declared search budget. Advanced-domain benchmarks make this
requirement more concrete. FormalProofBench tests private advanced-undergraduate
and graduate problems, MA-ProofBench separates undergraduate and
PhD-qualifying analysis, CAM-Bench evaluates context-dependent targets in
computational and applied mathematics, and Lean-IMO-Bench emphasizes
non-routine multi-step reasoning
~\citep{ravi2026formalproofbench,pu2026maproofbench,long2026cambench,
kung2026leap}. These benchmarks isolate different sources of difficulty and
should be treated as complementary stress tests, not combined into a single
leaderboard.

A credible data study should split by problem family rather than only by
individual theorem, remove near-duplicate statements and proof variants across
splits, and freeze the formal-library snapshot, retrieval index, and inference
budget. The same prover should then be compared with and without the proposed
synthetic or curated data. Alongside verified proof success, the study should
report semantic-audit outcomes, transfer across mathematical domains, and the
human effort required to repair proofs after library changes. A claimed gain is
not robust if it disappears after family-level deduplication, depends on access
to test-family artifacts, or does not survive the declared library version.
Releases should therefore separate source material, synthesized items,
filtering decisions, and downstream evaluation.
Figure~\ref{fig:provisional-failure-cases} maps five recurrent failure classes
to the distinct checks required to diagnose them.
\end{revisionblock}

\subsection{\revadd{Representing Relations Across Problems and Libraries}}
\label{subsec:deep_relationships}

\revadd{Many benchmark suites evaluate isolated theorem statements
~\citep{zheng2022minif2f,tsoukalas2024putnam}, whereas project-scale
formalization tasks inherit library dependencies and repository constraints
~\citep{xie2026mathlibpr}. This contrast motivates a research hypothesis:
representations linking informal concepts, formal declarations, and
tactic-level proof fragments may support retrieval and abstraction across
related tasks.}

\revadd{Candidate representations include proof-plan hierarchies
~\citep{jiang2023draftsketch,dong2024hierarchical,ren2025deepseekproverv2},
natural-language graphs of definitions, theorems, problems, and their reference
relations~\citep{bian2025automathkg}, and statement-level graphs linking
informal mathematics to formal-library declarations and dependencies
~\citep{kurgan2026theoremgraph}. Anti-unification supplies a complementary
mechanism for abstracting common structure from related terms or proofs
~\citep{cerna2023anti}. Barkeshli, Douglas, and Freedman's universal-proof and
structural-hypergraph proposal pushes the same question to corpus scale; it is a
conceptual research agenda rather than a reported capability result
~\citep{BarkeshliDouglasFreedman2026Structure}.}

\begin{figure}[h]
    \centering
    \includegraphics[width=0.98\linewidth]{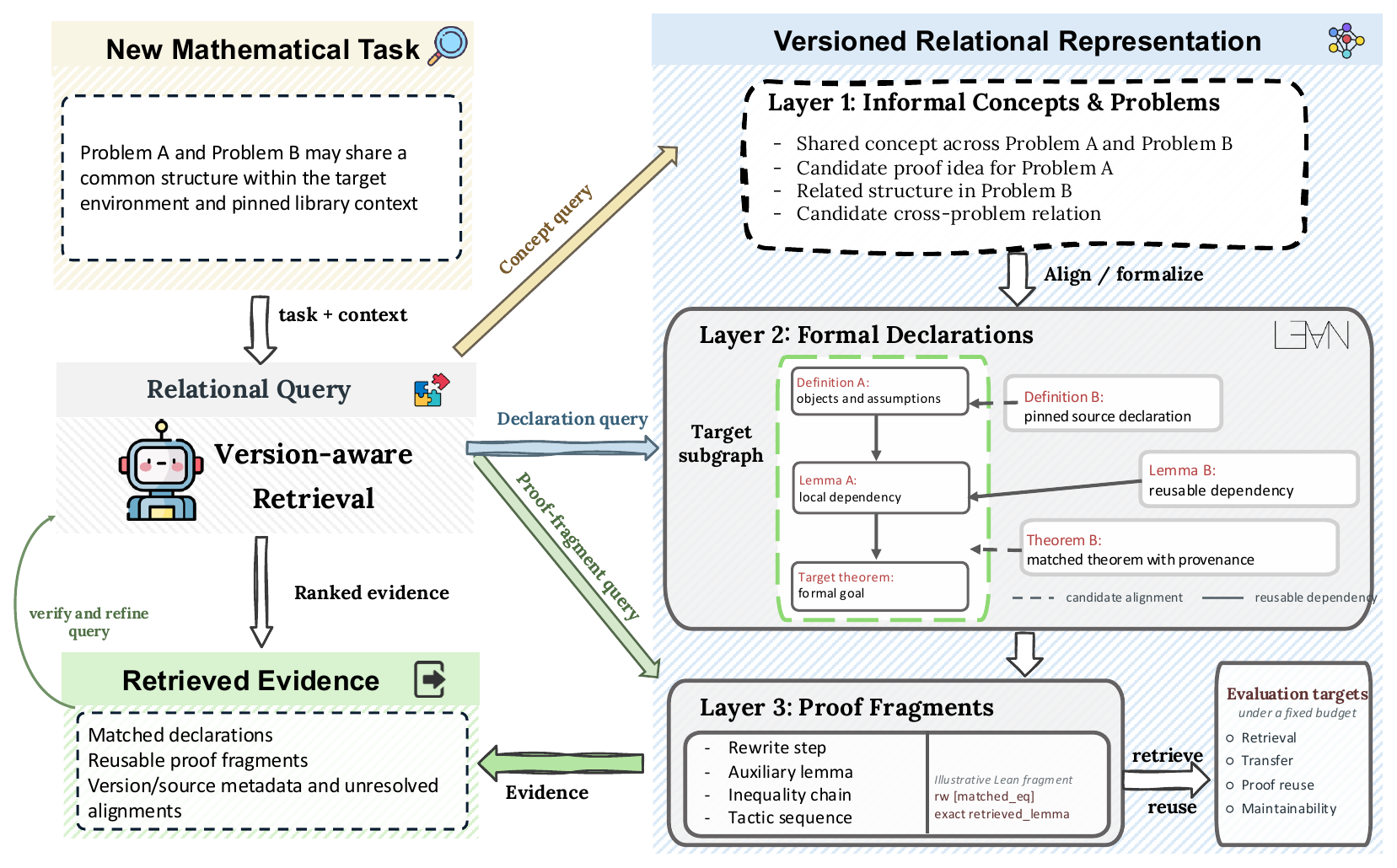}
    \caption{\textbf{Versioned relational representation for mathematical
    retrieval and reuse.} The diagram separates informal concepts and
    problems, formal declarations, and reusable proof fragments. Queries
    retrieve ranked evidence from a versioned store, while candidate
    alignments and reusable dependencies connect declarations across problems.
    Evaluation should measure retrieval, transfer, proof reuse, and
    maintainability under a fixed budget.}
    \label{fig:provisional-math-knowledge-graph}
\end{figure}

\revadd{This hypothesis can be tested with a versioned retrieval protocol.
Construct the representation from a fixed library snapshot and exclude entire
problem families from index construction and tuning. Evaluate first on those
held-out families in the same snapshot, then on a later library version in
which declarations, namespaces, or dependencies may have changed. Comparisons
should use the same candidate pool, prover, and search budget, with a lexical
retriever and the same prover without relational memory as controls. Reporting
should distinguish retrieval recall, provenance accuracy, proof reuse, and
maintenance cost. Provenance accuracy asks whether each retrieved artifact is
linked to the correct declaration, source version, and dependency closure;
proof reuse asks whether retrieved lemmas or fragments contribute to a
kernel-checked target proof; and maintenance cost records the expert time and
edits needed to restore validity after a library update. A representation has
not demonstrated transferable structure if its gain is confined to the source
snapshot, relies on train--test family overlap, or produces reusable-looking
fragments whose provenance or dependencies cannot be verified.
Figure~\ref{fig:provisional-math-knowledge-graph} instantiates the proposed
three-layer representation and its retrieval-and-reuse loop.}

\subsection{\revadd{Evaluating Conjecture and Concept Generation}}
\label{subsec:discovery}

\begin{revisionblock}
One objective in AI4Math is to support conjecture generation as well as proof.
Systems such as STP alternate theorem proposal and proof attempts to generate
self-supervised formal tasks~\citep{dong2025stp}. Formal validity does not by
itself establish that the resulting statements are useful or mathematically
novel.
\end{revisionblock}

\begin{revisionblock}
Evaluator-guided search broadens the candidate space but does not by itself
establish mathematical discovery. AlphaEvolve mutates executable candidates
under task-specific evaluators~\citep{novikov2025alphaevolve}; a separate
mathematical report describes improved bounds for selected Erd\H{o}s problems
~\citep{georgiev2025mathematical}. FormalEvolve applies archive-based
evolutionary search to autoformalization, whereas ProofEvolve evolves
kernel-verified proof DAGs and a persistent schema library
~\citep{lu2026formalevolve,ye2026proofevolve}. Because these systems optimize
different objects, direct comparison would require a common validity test,
prior-art protocol, compute budget, and candidate-selection rule. A stronger
test of abstraction is whether a proposed definition, representation, or
intermediate lemma simplifies previously separate tasks that were not used to
generate or score it.
\end{revisionblock}


\subsection{\revadd{External-Tool Integration}}
\label{subsec:external-tool}

\begin{revisionblock}
Proof assistants already interface with external automated provers and SMT
solvers, verified numerical methods, and computer algebra
systems~\citep{czajka2018hammer,ekici2017smtcoq,
mohamed2025lean,kellison2022verified,lewis2017leanmm}.
Tao's standalone Python proof assistant separately targets arguments involving
estimates~\citep{tao_proof_assistant_2024}.
The open challenge is auditable composition rather
than tool invocation alone. For every external call, a workflow should preserve
the versioned input, assumptions, encoding, output, and the status of any
certificate or proof reconstruction.

The evidence required depends on the tool. For satisfiable instances, a
returned model can often be checked against the encoding; for unsatisfiable
instances, independent evidence may require a proof log and reconstruction
~\citep{barbosa2022cvc5,schurr2021alethe}. Computer-algebra
outputs can depend on branch conventions, hidden assumptions, or numerical
approximation~\citep{wester1999review,stoutemyer2013errors,lewis2017leanmm}.
Under a fixed compute budget, evaluation should therefore report mathematical
success, certificate or reconstruction rate, tool-failure detection, and
robustness to altered assumptions. Gains from tool-integrated language models
and prover delegation remain protocol-specific
~\citep{gou2024tora,blanchette2016hammering,mihal2025leanhammer}. A sharper
target is end-to-end certificate recovery: how often external outputs can be
reconstructed inside the prover, whether failures are detected, and whether
results survive altered assumptions without silently enlarging the trusted
base.
\end{revisionblock}

\Needspace{8\baselineskip}
\subsection{\revadd{Human--AI Collaboration}}
\label{subsec:human-collaboration}

\begin{revisionblock}
Collaboration should be analyzed as a workflow rather than inferred from a
system label. Automated components should sit inside a documented process in
which problem specification, interpretation, attribution, and responsibility
remain explicit
~\citep{collins2024building,naskrecki2025mathematicaldiscovery}. In general
human--AI interaction, AI Chains makes prompt chains and intermediate outputs
visible for user inspection and control~\citep{wu2022ai}. Existing
formal-mathematics interfaces span in-editor tactic suggestions and premise retrieval
~\citep{song2025leancopilot,yang2023leandojo}, formalization conducted with
mathematicians~\citep{liu2026numina}, a living proof blueprint under continuous
human guidance~\citep{zhang2026mathcopilot}, and terminal-based monitoring and
steering of proof search~\citep{kripner2026openprover}.

These records illustrate distinct interaction designs rather than provide a
controlled comparison. A comparative study would need to fix the task,
interface, budget, and human role, then measure proof success together with
expert time, error detection, correction, readability, and maintainability.
Planning and
proof-refactoring systems provide candidate mechanisms for exposing
intermediate lemmas and higher-level structure
~\citep{jiang2023draftsketch,cao2025reviving,fu2026refactor,
gu2026proofoptimizer}.

At the research frontier, collaboration also depends on what can be inspected.
A closed workflow may release a manuscript or a checked formal artifact while
not releasing the model, prompts, retrieval corpus, intermediate trajectories,
failed branches, or internal tools. Such a release can support artifact review
without supporting replay of the discovery process or precise attribution of
individual steps. Reproducibility should therefore be stated at the level
actually achieved: rebuilding the released formal artifact, rerunning the
documented workflow, or independently reconstructing the mathematical
argument. These support different claims and should not be treated as
equivalent.

First Proof's third batch makes the inspectability problem measurable.
ProofCouncil met the organizers' correctness threshold on nine problems,
but received full exposition credit on four. The report also describes
a ProofCouncil harness loop that displaced a better argument and a MerLean
submission that relied on an obsolete definition
~\citep{abouzaid2026thirdbatch}. These observations motivate retaining
intermediate arguments, checking the active target after each revision,
and measuring expert reconstruction effort alongside final correctness.

Provenance is equally important in mixed human--machine work. A useful record
should connect source material, retrieved declarations, generated
intermediates, human corrections, and the final proof or exposition. This makes
it possible to locate semantic changes and distinguish reuse from new
contribution. The reporting checklist in
Section~\ref{sec:literature-map-2026} specifies a minimum record, but the open
research question is whether provenance can be captured during the workflow
without making expert collaboration impractical.

Resource concentration creates a separate evaluation problem. The OpenAI
Navier--Stokes record reports approximately 10,000 concurrent agents, 2.7
million exchanged messages, and 130 billion output tokens, while the research
model itself remained internal
~\citep{OpenAI2026NavierStokesAnnouncement}. This is evidence about one
workflow, not the prevalence of such resource use. It nevertheless shows why a
result obtained with private models, specialized infrastructure, or unusually
large compute should not be compared with an open assistant as if the model
name were the only changing variable. Collaboration studies should report the
resource envelope and include matched-budget or reduced-budget controls where
possible.

Avigad argues for a deliberately broader design space encompassing
formalization and digitization, curated libraries, proof assistants, symbolic
solvers, machine-learning-based pattern discovery, and new forms of interaction,
not only neural theorem proving. This is an agenda for tool and curriculum
design, not a capability estimate~\citep{avigad2026whatmathematics}.
\end{revisionblock}

\begin{revisionblock}
\Needspace{12\baselineskip}
\subsubsection{\revadd{Mathematicians' Perspectives}}
\label{subsec:mathematician-perspectives}

These public reflections illuminate questions of mathematical practice and
stewardship. The sample is purposive and spans essays, interviews, lectures,
and position pieces. It is not an estimate of community opinion or system
capability, and awards are noted only as context.

One theme is conditional and auditable capability. Tao frames future
capability as hypotheses parameterized by the task, field, tools, cost, human
supervision, and success criterion. He anticipates bottlenecks not only in
verification but also in exposition, publication, collective digestion, and
canonicalization~\citep{Tao2026MathematicsAgeAI}. Hairer turns
the same concern into writing practice through adversarial checking, source
tracing, rewriting in the author's own words, and disclosure when AI materially
affects the mathematical content~\citep{Hairer2026WritingAI}.

A second theme is that correctness does not exhaust understanding.
Williamson discusses whether AI will help or hinder mathematical
understanding~\citep{Williamson2026ICMPublicLecture}. Under his capability
hypothesis, Tao warns that optimizing benchmarkable achievements may pull
mathematical goals apart and that proof generation may outpace verification,
exposition, and community uptake~\citep{Tao2026MathematicsAgeAI}. Gowers
emphasizes the insight and communication that make a correct solution
intelligible, including discovery moves commonly omitted from finished proofs
~\citep{Gowers2026ChiefScientistInterview,Gowers2025MotivatedProofs}.
His proposed division between broad model knowledge
and branching and human pruning remains a time-stamped working hypothesis
~\citep{Gowers2026WhatSort}. Viazovska locates mathematical value in posing
meaningful questions and understanding their answers; a later report she
coauthored distinguishes kernel acceptance from readable, maintainable, and
reusable formal code
~\citep{ViazovskaInterview2022,HariharanEtAl2026SpherePacking}.

These concerns extend from individual understanding to stewardship. Gowers
questions categorical human-only credit rules while warning about erosion of
expertise and mathematical culture~\citep{Gowers2026Leiden}. Venkatesh
emphasizes conceptual language, mental representation, communication, and the
social processes by which a field assigns significance
~\citep{Venkatesh2024Automation,Venkatesh2025HumanMathematics,
LevinStrogatz2026ICM}. Scholze's Leiden Declaration endorsement likewise
stresses human understanding and the communal development of ideas
~\citep{Scholze2026LeidenEndorsement}. A joint \emph{Notices} position article
connects these priorities to community-owned infrastructure, transparent
reporting and attribution, and education centered on judgment
~\citep{CommelinEtAl2026Shaping}.

Recent first-person accounts also show how quickly practice can change.
In an interview published in December 2025, Duminil-Copin described AI as
useful mainly for writing and transcription. In August 2026, he reported
that frontier models surpassed him on some tasks. He also warns that rapid resolution of open
conjectures may bypass the failed attempts, tool-building, and apprenticeship
through which researchers and fields develop
~\citep{MilhazesFreitas2025DuminilCopinInterview,
DuminilCopin2026Care}. Among the IMU's four 2026 Fields Medalists
~\citep{IMU2026FieldsMedals}, Tsimerman reports using AI to accelerate routine
work while independently checking its outputs
~\citep{RobertsTsimerman2026FunFaster}. Deng's interview discusses AI-assisted
lemma proving and the expansion of mathematical
tools~\citep{Deng2026SAIRInterview}.

Sarnak's 2026 ``AlphaZero Test'' turns a broader disagreement into a bounded
experiment: ask a theorem prover to decide the Ramanujan conjecture from its
elementary statement while withholding outside data and theory. The article
proposes this test but reports no implementation; its predicted outcome remains
conjectural~\citep{Sarnak2026AlphaZeroTest}. Across these perspectives, the
recurring issue is not only whether a correct artifact can be produced, but how
it is explained, attributed, taught, reviewed, and integrated into mathematical
practice.
\end{revisionblock}

\section{Conclusion}
\label{sec:conclusion}

\begin{revisionblock}
AI4Math systems increasingly support mathematical research through retrieval,
formalization, proof search, tool use, and collaboration. Their contributions
remain tied to particular tasks and workflows. A checked proof establishes a
formal result under stated assumptions; fidelity to the intended mathematics,
novelty, significance, and human contribution require separate evidence.

Progress across the five research priorities depends on preserving mathematical
meaning and evidence as work moves between informal problems, formal targets,
external tools, and human judgment. Evaluation should account for generalization,
reproducibility, resource use, and human intervention. The reporting checklist
in Section~\ref{sec:literature-map-2026} provides a common basis for documenting
these factors. By evaluating versioned workflows and the mathematical
contributions they produce, the field can connect advances in proof generation
to reliable, reusable support for research.
\end{revisionblock}

\Needspace{10\baselineskip}
\begingroup
\setlength{\bibsep}{7.25pt plus 3pt minus 1.5pt}
\phantomsection
\pdfbookmark[0]{References}{pdf-references}
\bibliographystyle{arxiv_version}
\bibliography{custom,updates_2026,sources_extreme_2026,sources_multimedia_2026,sources_icm_multimedia_2026,sources_timeline_2026,sources_late_breaking_2026,sources_cutoff_2026,sources_cutoff_gap_2026,sources_mathematician_views_2026,sources_postcutoff_2026,sources_october_2026}
\endgroup

\end{document}